\documentclass[lettersize,journal]{IEEEtran}

\usepackage{graphicx}
\usepackage{amssymb,amsmath,epsfig}
\usepackage{algorithm}
\usepackage{algorithmic}
\usepackage{booktabs}
\usepackage{mathrsfs}
\usepackage{color}
\usepackage{textcomp}
\usepackage{bbding}
\usepackage{pifont}
\usepackage{wasysym}
\usepackage{amssymb}
\usepackage{subcaption}
\usepackage{cite}
\usepackage{amsmath,amssymb,amsfonts}
\usepackage{xcolor,colortbl}
\usepackage{overpic}

\usepackage{microtype}
\usepackage{multirow}
\usepackage{makecell}
\usepackage{hhline}
\usepackage[american]{babel}
\usepackage{cuted}
\usepackage{ragged2e}

\definecolor{lightgray}{gray}{.92}
\definecolor{tinygray}{gray}{.96}

\newcommand{\ie}{\textit{i}.\textit{e}.}
\newcommand{\eg}{\textit{e}.\textit{g}.}
\newcommand{\etc}{\textit{etc}}

\usepackage{multirow}
\usepackage{epsfig}
\usepackage{adjustbox}
\usepackage[
            breaklinks=true,colorlinks,
            citecolor=cyan,linkcolor=red,
            bookmarks=false]{hyperref}

\ifCLASSINFOpdf
\else
\fi
\def\BibTeX{{\rm B\kern-.05em{\sc i\kern-.025em b}\kern-.08em
    T\kern-.1667em\lower.7ex\hbox{E}\kern-.125emX}}

\usepackage[switch]{lineno}

\begin{document}

\title{FourierSR: A Fourier Token-based Plugin for Efficient Image Super-Resolution}

\author{Wenjie Li,
        Heng Guo\textsuperscript{$\ast$},
        Yuefeng Hou,
        and Zhanyu Ma
\IEEEcompsocitemizethanks{\IEEEcompsocthanksitem This work was supported by Beijing-Tianjin-Hebei Basic Research Funding Program No. F2024502017, National Natural Science Foundation of China (Grant No. 62472044, U24B20155, 62225601, U23B2052), Hebei Natural Science Foundation Project No. 242Q0101Z, Beijing Natural Science Foundation Project No. L242025, and in part by the Fundamental Research Funds for the Beijing University of Posts and Telecommunications under Grant 2025AI4S15. ($\ast$: Corresponding author) 
}
\IEEEcompsocitemizethanks{\IEEEcompsocthanksitem Wenjie Li, Heng Guo, and Zhanyu Ma are with the Pattern Recognition and Intelligent System Laboratory, School of Artificial Intelligence, Beijing University of Posts and Telecommunications (BUPT), Beijing 100080, China (e-mail: \{cswjli, guoheng, mazhanyu\}@bupt.edu.cn) .
}
\IEEEcompsocitemizethanks{\IEEEcompsocthanksitem Yuefeng Hou is with the School of Microelectronics, Tianjin University (TJU), Tianjin 300072, China (e-mail: houyuefeng@tju.edu.cn).
}
}

\markboth{IEEE TRANSACTIONS ON IMAGE PROCESSING}%
{Shell \MakeLowercase{\textit{et al.}}: A Sample Article Using IEEEtran.cls for IEEE Journals}

\maketitle

\begin{abstract}
Image super-resolution (SR) aims to recover low-resolution images to high-resolution images, where improving SR efficiency is a high-profile challenge. However, commonly used units in SR, like convolutions and window-based Transformers, have limited receptive fields, making it challenging to apply them to improve SR under extremely limited computational cost. To address this issue, inspired by modeling convolution theorem through token mix, we propose a Fourier token-based plugin called FourierSR to improve SR uniformly, which avoids the instability or inefficiency of existing token mix technologies when applied as plug-ins. Furthermore, compared to convolutions and windows-based Transformers, our FourierSR only utilizes Fourier transform and multiplication operations, greatly reducing complexity while having global receptive fields. Experiments show that our FourierSR as a plugin brings an average PSNR gain of 0.34dB for existing efficient SR methods on Manga109 test set at the scale of $\times 4$, while the average increase in the number of Params and FLOPs is only 0.6\% and 1.5\% of original sizes. Code link: \url{https://github.com/PRIS-CV/FourierSR}.
\end{abstract}



\IEEEpeerreviewmaketitle

\section{Introduction}
\IEEEPARstart{I}{MAGE} super-resolution (SR) aims to reconstruct a low-resolution (LR) image to a high-resolution (HR) image, which plays an important role in computer vision, such as visual quality improvement~\cite{jiang2025survey} and security surveillance~\cite{li2023survey}. To promote the application of SR on devices with limited computational resources, efficient SR has been continuously studied. However, existing methods face a trade-off between performance and computational cost. Therefore, a plug-and-play unit with the ability to efficiently enhance performance while consuming minimal computational costs is desired.


The challenge of the plug-and-play unit for efficient SR lies in balancing complexity and performance. Existing CNN-based methods utilize techniques such as feature distillation~\cite{hui2019lightweight} and wide activation~\cite{gao2022feature} to reduce model complexity, of which 3$\times$3 convolutions are widely used. However, compared to the local receptive fields from convolution, larger receptive fields are preferred for improving the performance of SR. Transformer can be one solution for obtaining global receptive fields, but it also introduces high computational costs. Existing Transformer-based efficient SR methods~\cite{liang2021swinir} reduce the complexity by dividing inputs into small windows. Due to limited connections between different windows, the receptive field of models is limited. Therefore, this paper aims to design an efficient SR plugin with low costs and global receptive fields.

\begin{figure}[t]
\begin{overpic}[width=0.99\linewidth]{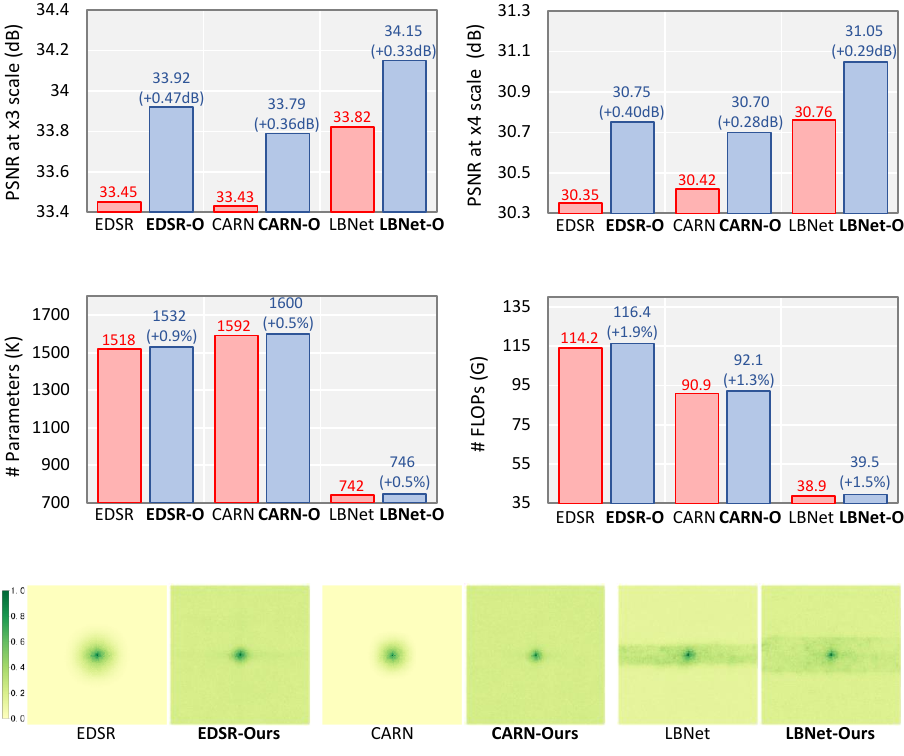}
\put(17.7,54.0){\color{black}{\fontsize{7pt}{1pt}\selectfont (a) Comparisons of PSNR on Manga109 \cite{matsui2017sketch} test set.}}
\put(7.7,21.8){\color{black}{\fontsize{7pt}{1pt}\selectfont (b) Comparisons of the number of Params and FLOPs at the scale of $\times 4$.}}
\put(8.2,-3.0){\color{black}{\fontsize{7pt}{1pt}\selectfont (c) Comparisons of the effective receptive field (ERF) \cite{luo2016understanding} visualization.}}
\end{overpic}
\vspace{5mm}
   \caption{(\textbf{Top}) The performance and efficiency improvement embedded with our FourierSR (-O), a plugin for improving existing efficient SR methods \cite{lim2017enhanced,ahn2018fast,gao2022lightweight}. (\textbf{Bottom}) With our FourierSR plugin, the receptive field of existing efficient SR methods \cite{lim2017enhanced,ahn2018fast,gao2022lightweight} can be extend effectively.
   }
\label{fig: teaser}
\vspace{-0mm}
\end{figure}

Inspired by performing token mixing in the Fourier domain to model global convolution via convolution theorem~\cite{oppenheim1999discrete} at low costs in image recognition and detection tasks~\cite{li2020fourier}, token mix can be one solution for designing efficient SR plugins, where token mix refers to exchange information contained in divided features of inputs. Despite the potential of applying token mix techniques to SR tasks, it is non-trivial to apply the token mix in efficient SR plugins where expensive costs or instability could be introduced. For example, AFNO~\cite{guibas2021efficient} utilizes a set of matrix multiplications to mix channel tokens in the Fourier domain, bringing computational costs to be too expensive for efficient SR. GFNet~\cite{liu2021swin} defines global filters the same size as inputs to perform token mix. However, in the SR task, the input size differs between training and testing, making it unstable to apply this approach consistently. AFFNet~\cite{huang2023adaptive} multiplies mixed tokens with Fourier features. However, the multiplication of complex tensors produces unstable values, leading to gradient explosion in SR training.

To construct improved Fourier tokens suitable for the efficient SR plugin, we propose FourierSR to boost off-the-shelf efficient SR methods. Similar to existing token mix methods, we simulate the global convolution based on the convolution theorem~\cite{oppenheim1999discrete}, which states that element-wise products in the frequency domain are equivalent to convolutions in the time domain. The key difference is that we find using defined local kernels to perform token mixing on the real and imaginary parts of Fourier features results in more robust and superior performance. Specifically, we extend defined local kernels to obtain simulated global kernels, which allows FourierSR to avoid the size inconsistency caused by defining global filters as kernels for SR training and testing in GFNet~\cite{liu2021swin}, and the gradient explosion resulting from defining Fourier features as kernels in AFFNet~\cite{huang2023adaptive}. Furthermore, we demonstrate that multiplying the real and imaginary parts of Fourier features with defined filters further results in enhanced global receptive fields. Additionally, such operation is lightweight, avoiding the slow inference and large number of parameters associated with heavy computational operations in AFNO~\cite{rao2021global} and FFC~\cite{chi2020fast}.

Due to its global receptive field and its absence of heavy computational operations, our FourierSR is ideally suited as a plug-and-play unit to enhance efficient SR methods. Experiments show our FourierSR as a plugin can be easily adapted to off-the-shelf efficient methods and improve their performance while avoiding the significant gain of computational cost. As shown in Fig.~\ref{fig: teaser} (a) and (b), our FourierSR improves PSNR by more than 0.34dB while increasing the average number of parameters by only 0.6\% and FLOPs by only 1.5\%, which benefits from our FourierSR's ability to expand the limited receptive field significantly of existing efficient CNN~\cite{lim2017enhanced,ahn2018fast} and windows-based Transformer~\cite{gao2022lightweight} methods, as shown in Fig.~\ref{fig: teaser} (c). Our contributions can be summarized as:

\begin{itemize}
    \item We propose a plugin called FourierSR, which improves off-the-shelf efficient SR by simulating global convolution based on the token mix and convolution theorem. 
    \item We propose a token mix scheme that uses local filters to modulate real and imaginary parts of Fourier features, efficiently and robustly achieving global receptive fields. 
    \item Experiments show FourierSR can be a plug-and-play unit that efficiently expands receptive fields of existing methods and thus improves SR quality with minimal costs.
\end{itemize}

\section{Related Work}

\subsection{Efficient SR Methods}
To reduce the high computational costs of general SR methods~\cite{li2018multi, zhang2018dcsr, fang2020soft, cai2023hipa, li2024systematic}, efficient SR has been developed recently. CARN~\cite{ahn2018fast} utilizes group convolution and recursive mechanisms to save SR costs. s-LWSR~\cite{li2020s} designs a lightweight network with a U-shaped backbone plus depth-separable convolution. IDN~\cite{hui2018fast} introduces the concept of information distillation to improve model speed while maintaining model performance. Subsequently, RFDN~\cite{liu2020residual}, FDIWN~\cite{gao2022feature}, and FIWHN~\cite{li2022efficient} further enhance the model's representation ability by improved information distillation. To obtain a larger receptive field, ShuffleMixer~\cite{sun2022shufflemixer} and CFSR~\cite{wu2024transforming} utilize separable large kernel convolution to efficiently model a wide range of receptive fields with minimal costs. SMFANet~\cite{zheng2025smfanet} further aggregates local and non-local features from large kernel convolution to compress models while improving SR performance. However, CNN-based SR methods still face the dilemma that it is difficult to further improve SR due to their limited receptive fields.

\begin{table}[t]
\tiny
\setlength\tabcolsep{1pt}
\centering
\vspace{0mm}
    \caption{Existing methods based on Fourier or token-mix, while capable of simulating non-local convolutions, are not suitable as plug-ins due to their poor stability or computational inefficiency.}
\label{tab: non_trivial}
\resizebox{0.48\textwidth}{!}{
\begin{tabular}{c|c|c}
\toprule
    \multicolumn{1}{c|}{\multirow{-1}{*}{Strategy}} 
    & \multicolumn{1}{c|}{\multirow{-1}{*}{Methods}}
    & \multicolumn{1}{c}{\multirow{-1}{*}{Why isn't suitable as plug-ins?}}

    \\ 
    \hline
    
    \makecell{Conducting convolution \\ in the Fourier domain}  & \makecell{\cite{chi2020fast}, \\ \cite{sinha2022nl}, \etc} & \makecell{Convolutions or linear layers \\ are computational inefficiency.} \\ 
    \hline
    
    \makecell{Defining global \\ filters for token mix}  & \makecell{\cite{rao2021global}, \\ \cite{tatsunami2024fft} , \etc \\ } & \makecell{Different size in train and test, \\ large number of Params.} \\ 
    \hline

    \makecell{Utilizing Fourier \\ features for token mix}  & \makecell{\cite{kong2023efficient}, \cite{huang2023adaptive} \\ \cite{lee2021fnet}, \etc } & \makecell{Prone to gradient \\ explosions during training.} \\ 
    \hline

    \makecell{A lot of matrix mul \\ in the Fourier domain}  & \makecell{\cite{li2020fourier}, \cite{guibas2021efficient}\\ \cite{yi2024fouriergnn} , \etc \\ } & \makecell{Slow inference speed.} \\ 
    \hline

    \makecell{windows-based mul \\ in the Fourier domain}  & \makecell{\cite{kong2023efficient}, \etc \\ } & \makecell{Limited ability of \\ model representation} \\ 
    \hline

    \bottomrule
\end{tabular}}
\end{table}
\hspace{-0mm}

To explore the global receptive field of models, SwinIR~\cite{liang2021swinir} first introduces the windows-based Transformer to capture long-range features. Then, A series of Transformer-based methods for efficient SR have been proposed. Specifically, ESRT~\cite{lu2022transformer} and LBNet~\cite{gao2022lightweight} split the vectors in the Transformer to reduce the costs of the Transformer. ELAN~\cite{zhang2022efficient} speeds up the Transformer's inference through self-attention share mechanisms. Onni-SR~\cite{wang2023omni}, WFEN~\cite{li2024efficient}, and DMNet~\cite{li2025dual} enhance the ability of the Transformer to extract features by expanding the range of pixel modeling. CFIN~\cite{li2023cross} improves the cross-scale interaction and contextual reasoning ability of the Transformer. NGSwin~\cite{choi2023n} attempts to solve the problem of difficulty in connecting information from different windows in the Transformer. SRFormer~\cite{zhou2023srformer} improves SR by permutation spatial information to channels. To reduce the complexity of the Transformer, MambaIR~\cite{guo2024mambair} first introduces the state space model to achieve long-range modeling with linear complexity. Unlike these methods that design complete networks, our method simulates global convolution by improved token mix in the Fourier domain, which can be an SR plugin to improve off-the-shelf efficient SR methods with minimal costs.

\begin{figure*}[ht]
\begin{overpic}[width=0.99\linewidth]{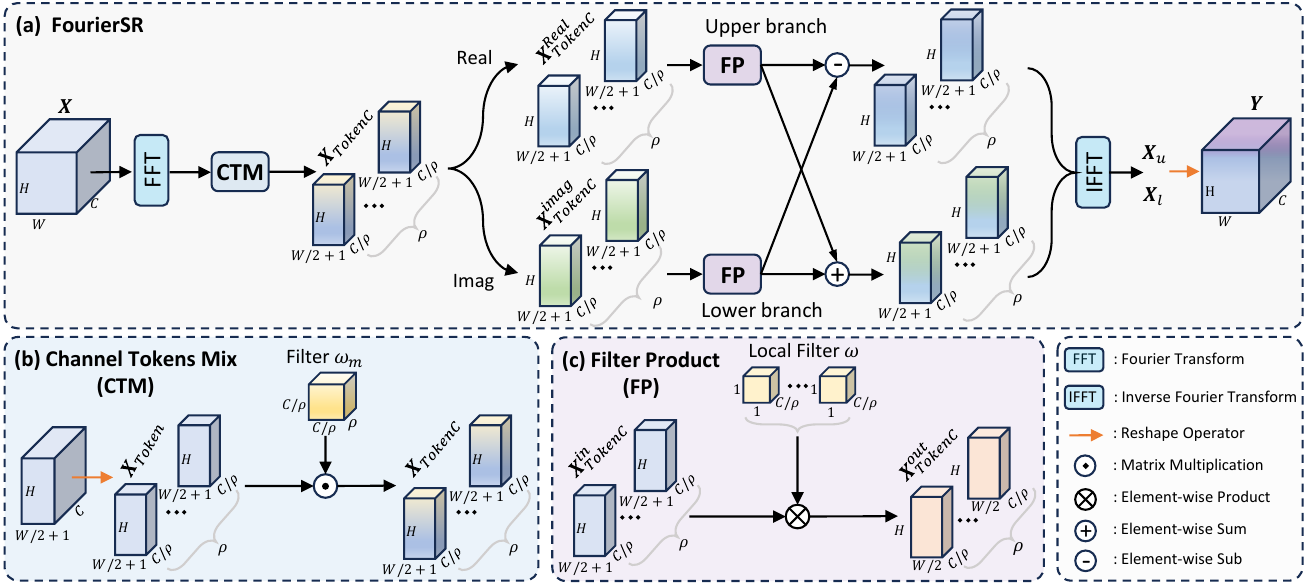}
\end{overpic}
\vspace{-0mm}
   \caption{Overview of (a) FourierSR, (b) Channel Tokens Mix (CTM) in FourierSR, (c) Filter Product (FP) in FourierSR.}
\label{fig: main}
\vspace{-0mm}
\end{figure*}

\subsection{Fourier-based and Token Mix-based Methods}
\textbf{Fourier-based Methods.} Fourier-based methods have grown significantly due to Fourier information can be used to improve the performance of models from a frequency analysis perspective. Specifically, in low-level tasks, NL-FFC~\cite{sinha2022nl} conducts convolutions in the Fourier domain to explore the relationship of frequency features. SFMNet~\cite{wang2023spatial} explores the frequency-spatial connections between the Fourier domain and spatial domain. FECNet~\cite{huang2022deep}, FSDGN~\cite{yu2022frequency}, FourLLIE~\cite{wang2023fourllie}, and FourierDiff~\cite{lv2024fourier} achieve precise restoration of image backgrounds by accurately reconstructing the amplitude and phase in the Fourier domain, enabling tasks such as defogging, low-light enhancement, and more. FreqMamba~\cite{zou2024freqmamba} improves the efficiency of models by connecting them to the Fourier transform for global degradation modeling. However, these methods utilize feature extraction modules in the Fourier domain, which is parametrically inefficient as plug-ins.

\textbf{Token Mix-based Methods.} Token mix-based methods have recently been introduced in high-level tasks, such as image recognition and detection
task~\cite{li2020fourier}, for exploring long-range features with less complexity. Specifically, FFC~\cite{chi2020fast} enhances representations through element-by-element addition and convolutions in the Fourier domain. GFNet~\cite{rao2021global} and DFFormer~\cite{tatsunami2024fft} explore global features by element-wise multiplication with global learnable filters. AFFNet~\cite{huang2023adaptive} performs semantic analysis in the Fourier domain to model large kernel convolution. FNO~\cite{li2020fourier}, AFNO~\cite{guibas2021efficient}, and FourierGNN~\cite{yi2024fouriergnn} simulate the global convolution through the matrix multiplication with learnable filters. However, as shown in the TABLE~\ref{tab: non_trivial}, the above Fourier-based or token-based methods are non-trivial in applying efficient SR due to their stability and efficiency. In this paper, we enhance off-the-shelf efficient SR methods with our improved Fourier tokens that apply to efficient SR.

\section{Method}
In this section, we first introduce preliminary knowledge, including token mix and the convolution theorem~\cite{oppenheim1999discrete} in Section~\ref{subsec:Preliminary}. Then, we introduce our FourierSR based on the above knowledge in Section~\ref{subsec:FourierSR}. Next, we analyze our strengths compared with related methods in Section~\ref{subsec:strength}. Finally, we introduce the location of our FourierSR insertion within existing efficient SR methods in Section~\ref{subsec:location}.


\subsection{Preliminary Knowledge} \label{subsec:Preliminary}
Our work aims to design a computationally efficient plugin with global receptive fields by token mix and convolution theorem. We will introduce the above concepts below.

\subsubsection{Token Mix}
Token mix has received attention for learning non-local features in high-level tasks. Before introducing the token mix, we first describe what a token is. For a feature tensor \hbox{${{\boldsymbol{X}}}$~$\in$~$\mathbb{R}^{H\times W \times C}$}, where \hbox{$H, W, C$} is height, width, and channel counts, respectively. This feature can be viewed as obtained by transforming a series of tokens \hbox{${{\boldsymbol{x}}}$}, where the shape of a token can be \hbox{$\mathbb{R}^{H\times W\times {(C \mathord{\left/
{\vphantom {C \rho }} \right.
\kern-\nulldelimiterspace} \rho) \times \rho}}$}, \hbox{$\mathbb{R}^{{(H \mathord{\left/
{\vphantom {H \rho }} \right.
\kern-\nulldelimiterspace} \rho )} \times \rho \times {(W \mathord{\left/
{\vphantom {W \rho }} \right.
\kern-\nulldelimiterspace} \rho )} \times \rho \times C}$}, or \hbox{$\mathbb{R}^{{(H \mathord{\left/
{\vphantom {H \rho }} \right.
\kern-\nulldelimiterspace} \rho )} \times \rho \times {(W \mathord{\left/
{\vphantom {W \rho }} \right.
\kern-\nulldelimiterspace} \rho )} \times \rho \times {(C \mathord{\left/
{\vphantom {C \rho }} \right.
\kern-\nulldelimiterspace} \rho )} \times \rho}$} \etc, where $\rho$ is the number of tokens in the $H$, $W$ or $C$ dimension. On this basis, the process of mixing non-local features of original tokens \hbox{${{\boldsymbol{x}}}$} to get updated tokens ${\boldsymbol{x}^u}$ can be described:
\begin{equation}
{{\boldsymbol{x}}^u} = \sum\limits_{i \in \mathcal{N}\left( {\boldsymbol{x}} \right)} {{{\omega ^{i \to u}} \odot \phi \left( {{{\boldsymbol{x}}^i}} \right)} } ,
\label{eq: token-mix}
\end{equation}
where \hbox{${\mathcal{N}\left( \boldsymbol{x} \right)}$} are token features of one or more dimensions of ${\boldsymbol{x}}$, \hbox{${\boldsymbol{x}^i}$} is a token in \hbox{${\mathcal{N}\left( \boldsymbol{x} \right)}$}. ${{\omega ^{i \to u}}}$ is the weight exchange or mix, $\odot$ is the matrix multiplication or element-wise product operator. $\phi $ is the feature embedding function.

\begin{figure*}[ht]
\begin{overpic}[width=0.99\linewidth]{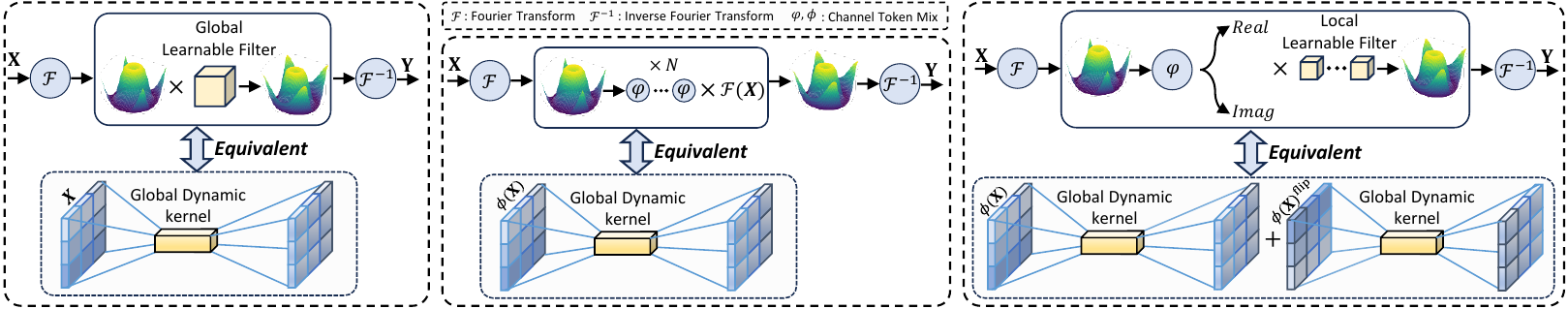}
\put(4.1,-2){\color{black}{\fontsize{8pt}{1pt}\selectfont (a) GFNet and DFFormer, \etc}}
\put(36.7,-2){\color{black}{\fontsize{8pt}{1pt}\selectfont (b) FNO and AFFNet, \etc}}
\put(74.7,-2){\color{black}{\fontsize{8pt}{1pt}\selectfont (c) \textbf{Our FourierSR}}}
\end{overpic}
\vspace{5.5mm}
   \caption{Comparisons of our FourierSR and existing Fourier token-based methods. (a) GFNet~\cite{rao2021global} and DFFormer~\cite{tatsunami2024fft} directly define global filters with the same size as inputs as kernels to simulate global convolutions. (b) FNO~\cite{li2020fourier}, FFTformer~\cite{kong2023efficient} and AFFNet~\cite{huang2023adaptive} use $N$ ($N \ge 8$) matrix multiplications for channel mix or define Fourier features as kernels to simulate global convolutions. (c) Our FourierSR extends local filters to global as global kernels to multiply with real and imaginary parts of Fourier features, which is equivalent to global convolutions of an input tensor plus its flipped.}
\label{fig: difference}
\vspace{-0mm}
\end{figure*}

\subsubsection{Convolution Theorem}
The time domain convolution theorem~\cite{oppenheim1999discrete}, in which convolutions in the time domain correspond to element-wise products in the frequency domain. This theorem can be formulated as:

\begin{equation}
\mathcal{F}\left[ {\boldsymbol{g}\left( t \right) * \boldsymbol{h}\left( t \right)} \right] = \mathcal{F}\left[ {\boldsymbol{g}\left( t \right)} \right] \times \mathcal{F}\left[ {\boldsymbol{h}\left( t \right)} \right] ,
\label{eq: convolution_theorem}
\end{equation}
where \hbox{$\mathcal{F}$} is the Fourier transform, \hbox{${\boldsymbol{g}\left( t \right)}$} and \hbox{${\boldsymbol{h}\left( t \right)}$} are time domain signals, \hbox{$ \times $} is a element-wise product, \hbox{$ * $} is a convolution. Next, we introduce some properties of the convolution theorem that will be used below. For a tensor \hbox{$\boldsymbol{X}$}, there are:
\begin{equation}
\mathcal{T}\left[ {\mathcal{F}\left( \boldsymbol{X} \right)} \right] = \mathcal{F}\left( \hat{\boldsymbol{X}} \right),\hat{\boldsymbol{X}} = \delta \left( \boldsymbol{X} \right) ,
\label{eq: transform}
\end{equation}
\begin{equation}
{\mathcal{F}^{\scriptscriptstyle -1}}\!\left[ {\mathcal{F}\left( {{\boldsymbol{X}} \pm {\hat{\boldsymbol{X}}}} \right)} \right] = {\mathcal{F}^{\scriptscriptstyle -1}}\!\left[ {\mathcal{F}\left( {{\boldsymbol{X}}} \right)} \right] \pm {\mathcal{F}^{\scriptscriptstyle -1}}\!\left[ {\mathcal{F}\left( {{\hat{\boldsymbol{X}}}} \right)} \right] ,
\label{eq: distributive_law}
\end{equation}
\begin{equation}
{\mathcal{F}^{\scriptscriptstyle -1}}\left\{ {\mathcal{F}\left[ {\boldsymbol{g}\left( t \right)} \right]} \right.\left. {\!\! \times \mathcal{F}\left[ {\boldsymbol{h}\left( t \right)} \right]} \right\} = \boldsymbol{g}\left( t \right)*\boldsymbol{h}\left( t \right) ,
\label{eq: Convolution_Theorem}
\end{equation}
where \hbox{${\mathcal{F}^{\scriptscriptstyle -1}}$} is an inverse Fourier transform, \hbox{$\mathcal{T}$} is a series of transformations, \hbox{$\hat{\boldsymbol{X}}$} is obtained after not changing the tensor size token mix \hbox{$\delta $} on \hbox{$\boldsymbol{X}$}. In addition, for the real and imaginary parts of Fourier features, according to deformations of the convolution theorem, there are:
\begin{equation}
{\mathcal{F}^{\scriptscriptstyle -1}}\left[ {Real\left( {\mathcal{F}\left( \boldsymbol{X} \right)} \right)} \right] = \frac{1}{2}\boldsymbol{X} + \frac{1}{2}{\boldsymbol{X}^{flip}} ,
\label{eq: real}
\end{equation}
\begin{equation}
{\mathcal{F}^{\scriptscriptstyle -1}}\left[ {Imag\left( {\mathcal{F}\left( \boldsymbol{X} \right)} \right)} \right] = \frac{1}{2}\boldsymbol{X} - \frac{1}{2}{\boldsymbol{X}^{flip}} ,
\label{eq: imag}
\end{equation}
where $Real$ and $Imag$ are real and imaginary parts of \hbox{${\mathcal{F}\left( \boldsymbol{X} \right)}$}, respectively. \hbox{${\boldsymbol{X}^{flip}}$} is the flip of tensor \hbox{${\boldsymbol{X}}$}.

\subsection{FourierSR} \label{subsec:FourierSR}
This subsection aims to prove mathematically that our FourierSR is equivalent to global dynamic convolutions.

\subsubsection{Filter Product (FP)} First, we need to describe \textbf{why local filters can be converted to global filters} during element products and with abilities to adjust network weights dynamically, which is crucial for proofing that our FourierSR can be equated to global dynamic convolutions.

For a tensor \hbox{$\boldsymbol{X}_{TokenC}^{in}$~$\in$~$\mathbb{R}^{\rho ,{C \mathord{\left/
{\vphantom {C \rho }} \right.
\kern-\nulldelimiterspace} \rho },H,\left( {{W \mathord{\left/
 {\vphantom {W 2}} \right.
 \kern-\nulldelimiterspace} 2} + 1} \right)}$} and a local filter \hbox{${{\omega }}$~$\in$~$\mathbb{R}^{\rho,{C \mathord{\left/
{\vphantom {C \rho }} \right.
\kern-\nulldelimiterspace} \rho }, 1, 1}$}, as shown in Fig.~\ref{fig: main} (b), when \hbox{${{\boldsymbol{X}_{TokenC}^{in}}}$} and \hbox{${{\omega }}$} do element-wise products, the broadcast mechanism of torch firstly expands \hbox{${{\omega }}$} into a global filter of the same size as \hbox{${{\boldsymbol{X}_{TokenC}^{in}}}$}, Then, \hbox{${{\omega }}$} as a learnable parameter can convert \hbox{$\boldsymbol{X}_{TokenC}^{in}$} into a learnable parameter to manage its weight optimization during training dynamically. Meanwhile, the number of params of this filter is still equal to the local filter rather than the global filter. The process of broadcast mechanism and obtaining \hbox{$\boldsymbol{X}_{TokenC}^{in}$~$\in$~$\mathbb{R}^{\rho ,{C \mathord{\left/
{\vphantom {C \rho }} \right.
\kern-\nulldelimiterspace} \rho },H,\left( {{W \mathord{\left/
 {\vphantom {W 2}} \right.
 \kern-\nulldelimiterspace} 2} + 1} \right)}$} with dynamical weights is:
\begin{equation}
\omega  \in \mathbb{R}^{\left[ {\rho ,{C \mathord{\left/
 {\vphantom {C \rho }} \right.
 \kern-\nulldelimiterspace} \rho },1,1} \right]} \to \omega  \in \mathbb{R}^{\left[ {\rho ,{C \mathord{\left/
 {\vphantom {C \rho }} \right.
 \kern-\nulldelimiterspace} \rho },H,\left( {{W \mathord{\left/
 {\vphantom {W 2}} \right.
 \kern-\nulldelimiterspace} 2} + 1} \right)} \right]},
 \label{eq: local_global}
\end{equation}
\begin{equation}
\boldsymbol{X}_{TokenC}^{out} = \omega  \in \mathbb{R}^{\left[ {\rho ,{C \mathord{\left/
 {\vphantom {C \rho }} \right.
 \kern-\nulldelimiterspace} \rho },H,\left( {{W \mathord{\left/
 {\vphantom {W 2}} \right.
 \kern-\nulldelimiterspace} 2} + 1} \right)} \right]} \times \boldsymbol{X}_{TokenC}^{in},
  \label{eq: learnable}
\end{equation}
therefore, local filters with light computational costs can be viewed as global filters in FP and make networks capable of adjusting weights dynamically in our FourierSR. Next, we will prove that our FourierSR is equal to the global dynamic convolution based on this.

\begin{table}[t]
\tiny
\setlength\tabcolsep{1pt}
\centering
\vspace{0mm}
    \caption{Complexity comparison between convolution (Conv), windows-based Transformer (WTrans)~\cite{liu2021swin}, existing token mix-based methods, \ie, FFC~\cite{chi2020fast}, GFNet~\cite{rao2021global}, AFNO~\cite{guibas2021efficient}, AFFNet~\cite{huang2023adaptive}, and our FourierSR. Our method has the smallest Params counts and the second smallest FLOPs counts. $k$: Convolution kernel size. $C$: The number of input and output channels. $H,W$: Height and width of input images. $M$: windows size in W-Trans. $\rho $: Tokens counts.}
\vspace{-0mm}
\label{tab: differnet}
\resizebox{0.495\textwidth}{!}{
\begin{tabular}{l|c|c}
\toprule
    \multicolumn{1}{l|}{\multirow{-1}{*}{Methods}}
    & \multicolumn{1}{c|}{\multirow{-1}{*}{FLOPs}}
    & \multicolumn{1}{c}{\multirow{-1}{*}{Parameters}}

    \\ 
    \hline
     Conv  & \hbox{${k^2}{C^2}HW$} & \hbox{${k^2}{C^2}$}\\ 
    WTrans~\cite{liu2021swin}  & \hbox{$4{C^2}HW + 2{M^2}CHW$} & \hbox{$4{C^2}$}\\
    FFC~\cite{chi2020fast}  & \hbox{${k^2}{C^2}HW\!+\!2CHW{\log _2}HW$} & \hbox{${k^2}{C^2}$}\\ 
    GFNet~\cite{rao2021global}  & \hbox{$CHW + 2CHW{\log _2}HW$} & \hbox{$CHW$}\\ 
    AFNO~\cite{guibas2021efficient}  & \hbox{${{8{C^2}HW} \mathord{\left/
 {\vphantom {{8{C^2}HW} \!\!\!\rho\!\! }} \right.
 \kern-\nulldelimiterspace} \rho } \!+\! 2CHW{\log _2}HW$} & \hbox{$(1\!+\!4\!/\!\rho ){C^2}\!\!+\!\!4C$} \\
    AFFNet~\cite{huang2023adaptive}  & \hbox{${{8{C^2}HW} \mathord{\left/
 {\vphantom {{8{C^2}HW} \!\!\!\rho\!\! }} \right.
 \kern-\nulldelimiterspace} \rho } \!+\! 2CHW{\log _2}HW$} & \hbox{$(1\!+\!4\!/\!\rho ){C^2}\!\!+\!\!4C$} \\
   \textbf{Ours}  & \hbox{${{{C^2}HW} \mathord{\left/
 {\vphantom {{8{C^2}HW} \rho }} \right.
 \kern-\nulldelimiterspace} \rho } \!+\! 2CHW{\log _2}HW$} & \hbox{$(6+\rho )C$} \\

    \bottomrule
\end{tabular}}
\end{table}
\hspace{-0mm}

\subsubsection{Mathematical Proof} As shown in Fig.~\ref{fig: main} (a), for an input tensor \hbox{${{\boldsymbol{X}}}$~$\in$~$\mathbb{R}^{C\times H \times W}$}, we first perform the Fourier transform on $\boldsymbol{X}$ and then utilize a reshape operator as the feature embedding unit $\phi $ to get Fourier tokens \hbox{${{\boldsymbol{X}_{Token}}}$~$\in$~$\mathbb{R}^{\rho ,{C \mathord{\left/
{\vphantom {C \rho }} \right.
\kern-\nulldelimiterspace} \rho },H,\left( {{W \mathord{\left/
 {\vphantom {W 2}} \right.
 \kern-\nulldelimiterspace} 2} + 1} \right)}$}. Next, we perform the channel tokens mix using matrix multiplication with our defined complex filter \hbox{${{{\omega}_{m} }}$~$\in$~$\mathbb{R}^{\rho,{C \mathord{\left/
{\vphantom {C \rho }} \right.
\kern-\nulldelimiterspace} \rho },{C \mathord{\left/
{\vphantom {C \rho }} \right.
\kern-\nulldelimiterspace} \rho }}$} to get a channel-mixed token \hbox{${{\boldsymbol{X}_{TokenC}}}$}. Combined with Eq.(\ref{eq: transform}), this process can be described and equated as: 
\begin{equation}
{\boldsymbol{X}_{TokenC}} =  {{\omega}_{m} \odot \phi \left( {\mathcal{F}\left( \boldsymbol{X} \right)} \right)}  = {\mathcal{F}\left( \hat{\boldsymbol{X}} \right)},
\end{equation}
where \hbox{$\hat{\boldsymbol{X}} = \delta \left( \boldsymbol{X} \right)$}, \hbox{$\hat{\boldsymbol{X}}$~$\in$~$\mathbb{R}^{\rho ,{C \mathord{\left/
{\vphantom {C \rho }} \right.
\kern-\nulldelimiterspace} \rho },H,\left( {{W \mathord{\left/
 {\vphantom {W 2}} \right.
\kern-\nulldelimiterspace} 2} + 1} \right)}$}, $\delta$ is a random channel token mix, $\odot$ is a matrix multiplication operator, $\mathcal{F}$ and $\mathcal{F}^{-1}$ have same meaning with Eq.(\ref{eq: transform}). Then, we perform the spatial token mix on height and width dimensions using our defined local filter $\omega $. For the upper branch in Fig.~\ref{fig: main} (a), combined with Eq.(\ref{eq: distributive_law}), the output \hbox{${\boldsymbol{X}_u}$} can be described and equated as:
\begin{equation}
\small
\begin{aligned}
{\boldsymbol{X}_u} \!= & {\mathcal{F}^{\scriptscriptstyle -1}}\!\left[ {\omega  \!\times\! Real\left( {\mathcal{F}\left( \hat{\boldsymbol{X}} \right)} \right) \!-\! \omega  \!\times\! Imag\left( {\mathcal{F}\left( \hat{\boldsymbol{X}} \right)} \right)} \right] \\
\!\!= \!\!&\ {\mathcal{F}^{\scriptscriptstyle -1}}\!\left[ {\omega  \!\times\! Real\!\left(\! {\mathcal{F}\left(\! \hat{\!\boldsymbol{X}} \!\right)} \!\right)} \!\right] \!\!-\! {\mathcal{F}^{\scriptscriptstyle -1}}\!\left[ {\omega  \!\times\! Imag\left(\! {\mathcal{F}\left(\! \hat{\boldsymbol{X}} \!\right)} \!\right)} \!\right]
\end{aligned}
\end{equation}
combined with Eq.(\ref{eq: local_global}), the size of \hbox{$\omega $~$\in$~$\mathbb{R}^{\rho ,{C \mathord{\left/
{\vphantom {C \rho }} \right.
\kern-\nulldelimiterspace} \rho },1,1}$} at this moment after broadcasting is same as \hbox{$Real\left( {\mathcal{F}\left( \hat{\boldsymbol{X}} \right)} \right)$}, \hbox{$Imag\left( {\mathcal{F}\left( \hat{\boldsymbol{X}} \right)} \right)$}, and \hbox{${\mathcal{F}\left( \hat{\boldsymbol{X}} \right)}$~$\in$~$\mathbb{R}^{\rho ,{C \mathord{\left/
{\vphantom {C \rho }} \right.
\kern-\nulldelimiterspace} \rho },H,\left( {{W \mathord{\left/
 {\vphantom {W 2}} \right.
 \kern-\nulldelimiterspace} 2} + 1} \right)}$}. Then, combined with Eq.(\ref{eq: Convolution_Theorem}), ${\boldsymbol{X}_u}$ can be further described as: 
\begin{equation}
\small
\begin{aligned}
{\boldsymbol{X}_u} \!= &\ {\mathcal{F}^{\scriptscriptstyle -1}}\!\left( \omega  \right)\!*\!{\mathcal{F}^{\scriptscriptstyle -1}}\!\left[ {Real\!\left( {\mathcal{F}\!\left( \hat{\boldsymbol{X}} \right)} \right)} \right] \\& -\! {\mathcal{F}^{\scriptscriptstyle -1}}\!\left( \omega  \right)\!*\!{\mathcal{F}^{\scriptscriptstyle -1}}\!\left[ {Imag\!\left( {\mathcal{F}\!\left( \hat{\boldsymbol{X}} \right)} \right)} \right],
\end{aligned}
\end{equation}
combined with Eq.(\ref{eq: real}) and Eq.(\ref{eq: imag}), we have:
\begin{equation}
\small
\begin{aligned}
{\boldsymbol{X}_u} \!\!=\! &\ {\mathcal{F}^{\scriptscriptstyle -1}}\!\left( \omega  \right)\!*\!\left[ {\frac{\hat{\boldsymbol{X}}}{2} \!+\! \frac{\hat{\boldsymbol{X}}^{flip}}{2}} \right] \!\!-\! {\mathcal{F}^{\scriptscriptstyle -1}}\!\left( \omega  \right)\!*\!\left[ {\frac{\hat{\boldsymbol{X}}}{2} \!-\! \frac{\hat{\boldsymbol{X}}^{flip}}{2}} \right],
\end{aligned}
\end{equation}
since convolutions satisfy the distributive law, the above equation can be equated to:
\begin{equation}
\small
\begin{aligned}
{\boldsymbol{X}_u} = &\ {\mathcal{F}^{\scriptscriptstyle -1}}\!\left( \omega  \right)\!*\!\left[ {\frac{\hat{\boldsymbol{X}}}{2} \!+\! \frac{\hat{\boldsymbol{X}}^{flip}}{2} \!-\! \frac{\hat{\boldsymbol{X}}}{2} \!+\! \frac{\hat{\boldsymbol{X}}^{flip}}{2}} \right] \\
= &\ {\mathcal{F}^{\scriptscriptstyle -1}}\!\left( \omega  \right)\!*\!{\hat{\boldsymbol{X}}^{flip}},
\label{eq: 14}
\end{aligned}
\end{equation}
since \hbox{$\omega$} is extended on spatial dimensions of the broadcast mechanism, the \hbox{$\omega$} now has the same size with \hbox{$\mathcal{F}\left( \hat{\boldsymbol{X}}  \right)$} and \hbox{$\mathcal{F}\left( \hat{\boldsymbol{X}}^{flip}  \right)$}, \hbox{${\mathcal{F}^{\scriptscriptstyle -1}}\left( \omega  \right)$} has the same size with 
\hbox{${{\hat{\boldsymbol{X}}}}, \hat{\boldsymbol{X}}^{flip}$~$\in$~$\mathbb{R}^{C\times H \times W}$}, which can be viewed a convolution kernel with global receptive field. According to the description in Eq.(\ref{eq: learnable}), \hbox{${\mathcal{F}^{\scriptscriptstyle -1}}\left( \omega  \right)$} can convert the tensor to a learnable parameter to dynamically adjust its weight. Therefore, \hbox{${\boldsymbol{X}_u}$~$\in$~$\mathbb{R}^{C\times H \times W}$} can be viewed as the result obtained by utilizing a convolution with a global receptive field kernel.

\begin{figure}[t]
\begin{overpic}[width=0.99\linewidth]{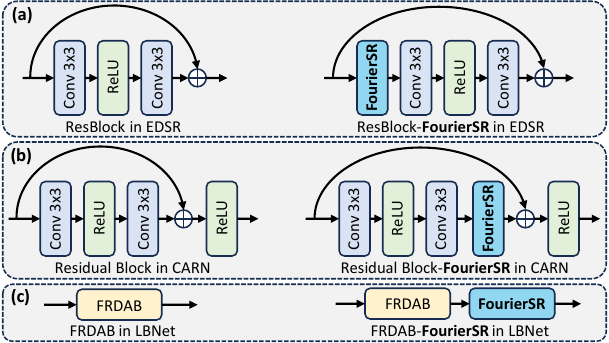}
\end{overpic}
\vspace{-0mm}
   \caption{Overview of positions where our FourierSR is inserted into methods listed in TABLE~\ref{tab: performance}. (a): EDSR~\cite{lim2017enhanced} and EDSR-Ours, (b): CARN~\cite{ahn2018fast} and CARN-Ours, (c): LBNet~\cite{gao2022lightweight} and LBNet-Ours.}
\label{fig: insert_table}
\vspace{-0mm}
\end{figure}

Similarly, combined with Eq.(\ref{eq: distributive_law}), Eq.(\ref{eq: Convolution_Theorem}), Eq.(\ref{eq: real}), and Eq.(\ref{eq: imag}), the output \hbox{${\boldsymbol{X}_l}$} of lower branch can be described as:
\begin{equation}
\small
\begin{aligned}
{\boldsymbol{X}_l} \!= & {\mathcal{F}^{\scriptscriptstyle -1}}\!\left[ {\omega  \!\times\! Real\left( {\mathcal{F}\left( \hat{\boldsymbol{X}} \right)} \right) \!+\! \omega  \!\times\! Imag\left( {\mathcal{F}\left( \hat{\boldsymbol{X}} \right)} \right)} \right] \\
= &\  ...     \\
= &\ {\mathcal{F}^{\scriptscriptstyle -1}}\!\left( \omega  \right)\!*\!\left[ {\frac{\hat{\boldsymbol{X}}}{2} \!+\! \frac{\hat{\boldsymbol{X}}^{flip}}{2} \!+\! \frac{\hat{\boldsymbol{X}}}{2} \!-\! \frac{\hat{\boldsymbol{X}}^{flip}}{2}} \right] \\
= &\ {\mathcal{F}^{\scriptscriptstyle -1}}\!\left( \omega  \right)\!*\!\hat{{\boldsymbol{X}}},
\end{aligned}
\end{equation}
similar with \hbox{${\boldsymbol{X}_u}$} in Eq.(\ref{eq: 14}), \hbox{${\mathcal{F}^{\scriptscriptstyle -1}}\left( \omega  \right)$} now has the same size with 
\hbox{${{\hat{\boldsymbol{X}}}}$~$\in$~$\mathbb{R}^{C\times H \times W}$}, \hbox{${\boldsymbol{X}_l}$~$\in$~$\mathbb{R}^{C\times H \times W}$} can be viewed as the result obtained by a global convolution to \hbox{${\hat{\boldsymbol{X}}}$}. Filters \hbox{$\omega$} are share-weighted with the upper branch. Finally, \hbox{${\boldsymbol{X}_u}$} and \hbox{${\boldsymbol{X}_l}$} obtained from the upper and lower branches are fused and reshaped to get the final output \hbox{${\boldsymbol{Y}}$}. Since \hbox{${\hat{\boldsymbol{X}}}$} and \hbox{${\hat{\boldsymbol{X}}^{flip}}$} are obtained by channel token mix, the final result \hbox{${\boldsymbol{Y}}$} can be regarded as obtained by token mix of channel features plus convolution with a global receptive field kernel on inputs after data enhancement, leading our FourierSR with a strong ability to capture long-range features interaction.

\begin{figure}[t]
\begin{overpic}[width=0.99\linewidth]{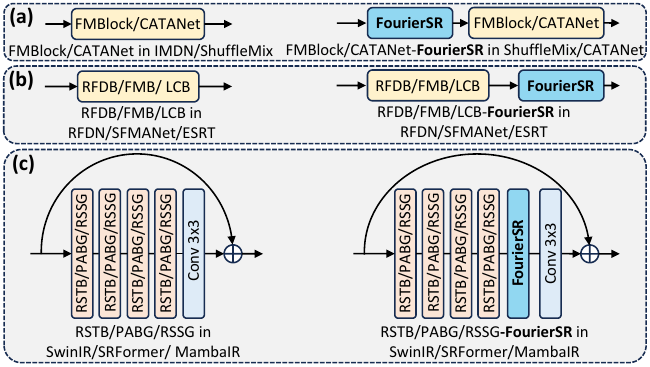}
\end{overpic}
\vspace{-0mm}
   \caption{Overview of positions where our FourierSR is inserted into methods listed in Fig~\ref{fig: more_psnr_lines}. (a): ShuffleMixer~\cite{sun2022shufflemixer}, CATANet~\cite{liu2025catanet} and (ShuffleMixer, CATANet)-Ours, (b): RFDN~\cite{liu2020residual}, SMFANet~\cite{zheng2025smfanet}, ESRT~\cite{lu2022transformer} and (RFDN, SMFANet, ESRT)-Ours, (c): SwinIR~\cite{liang2021swinir}, SRFormer~\cite{zhou2023srformer}, MambaIR~\cite{guo2024mambair} and (SwinIR, SRFormer, MambaIR)-Ours.}
\label{fig: insert_figure}
\vspace{-0mm}
\end{figure}


\begin{table*}[hbtp]
\tiny
\setlength\tabcolsep{2pt}
\centering
\vspace{0mm}
\caption{Quantitative evaluation of our method and existing efficient SR methods, where FLOPs is calculated on an upsampled image with a spatial size of 1280$\times$720. The best results are emphasized in \textbf{bold}.}
\vspace{-0mm}
\label{tab: performance}
\resizebox{0.98\textwidth}{!}{
\begin{tabular}{c|l|l|l|c|c|c|c|c}
\toprule
& & & & \multicolumn{1}{c|}{Set5 \cite{bevilacqua2012low}} 
& \multicolumn{1}{c|}{Set14 \cite{zeyde2012single}} 
& \multicolumn{1}{c|}{BSDS100 \cite{martin2001database}} 
& \multicolumn{1}{c|}{Urban100 \cite{huang2015single}} 
& \multicolumn{1}{c}{Manga109 \cite{matsui2017sketch}} 
\\ 
\cmidrule{5-9}
    \multicolumn{1}{c|}{\multirow{-2}{*}{Scale}}
    & \multicolumn{1}{l|}{\multirow{-2}{*}{Methods}}  
    & \multicolumn{1}{l|}{\multirow{-2}{*}{Params{\color[HTML]{369DA2} $\downarrow$}}}
    & \multicolumn{1}{l|}{\multirow{-2}{*}{FLOPs{\color[HTML]{369DA2} $\downarrow$}}}
    & PSNR{\color[HTML]{369DA2} }/SSIM{\color[HTML]{369DA2}}
    & PSNR{\color[HTML]{369DA2} }/SSIM{\color[HTML]{369DA2}}
    & PSNR{\color[HTML]{369DA2} }/SSIM{\color[HTML]{369DA2}}
    & PSNR{\color[HTML]{369DA2} }/SSIM{\color[HTML]{369DA2}}  
    & PSNR{\color[HTML]{369DA2} }/SSIM{\color[HTML]{369DA2}}
    \\ 
    \hline
     &  EDSR~\cite{lim2017enhanced}               & 1370K   &316.2G & 37.99/0.9604        &33.59/0.9175            & 32.16/0.8994    & 31.98/0.9272   & 38.55/0.9769    \\
    \rowcolor{gray!10}
    \cellcolor{white} &  \textbf{EDSR-Ours}            & 1384K   &326.5G  &  \textbf{38.07/0.9608}  & \textbf{33.66/0.9182}   & \textbf{32.19/0.8998}    & \textbf{32.27/0.9290}    & \textbf{38.89/0.9775}  \\
    &  \multicolumn{1}{l|}{ {\color[HTML]{369DA2} Gain}}               & \multicolumn{1}{l|}{ {\color[HTML]{369DA2} +14K (1.0\%)}}   &\multicolumn{1}{l|}{ {\color[HTML]{369DA2} +10.3G (3.3\%)}} & \multicolumn{1}{c|}{ {\color[HTML]{369DA2} +0.08/0.0004}}        &\multicolumn{1}{c|}{{\color[HTML]{369DA2} +0.07/0.0007}}            & \multicolumn{1}{c|}{ {\color[HTML]{369DA2} +0.03/0.0004}}    & \multicolumn{1}{c|}{ {\color[HTML]{369DA2} +0.29/0.0018}}   & \multicolumn{1}{c}{ {\color[HTML]{369DA2} +0.34/0.0006}}    \\
    \cmidrule{2-9}
    &  CARN~\cite{ahn2018fast}               & 1592K   &222.8G & 37.76/0.9590       & 33.52/0.9166           & 32.09/0.8978    & 31.92/0.9256     & 38.36/0.9765  \\
    \rowcolor{gray!10}  \cellcolor{white}
    $\times 2$
    &  \textbf{CARN-Ours}            &1600K     &228.6G  & \textbf{38.03/0.9605}       & \textbf{33.65/0.9179}           & \textbf{32.15/0.8993}    & \textbf{32.10/0.9276}     & \textbf{38.74/0.9771}  \\
    &  \multicolumn{1}{l|}{ {\color[HTML]{369DA2} Gain}}               & \multicolumn{1}{l|}{ {\color[HTML]{369DA2} +8K (0.5\%)}}   &\multicolumn{1}{l|}{ {\color[HTML]{369DA2} +5.8G (2.6\%)}} & \multicolumn{1}{c|}{ {\color[HTML]{369DA2} +0.27/0.0015}}        &\multicolumn{1}{c|}{{\color[HTML]{369DA2} +0.07/0.0013}}            & \multicolumn{1}{c|}{ {\color[HTML]{369DA2} +0.06/0.0015}}    & \multicolumn{1}{c|}{ {\color[HTML]{369DA2} +0.18/0.0020}}   & \multicolumn{1}{c}{ {\color[HTML]{369DA2} +0.38/0.0006}}    \\
    \cmidrule{2-9}
    &  LBNet~\cite{gao2022lightweight}               & 731K   &153.2G  & 38.05/0.9607       & 33.65/0.9177           & 32.16/0.8994    & 32.30/0.9291     & 38.88/0.9775    \\
    \rowcolor{gray!10}  \cellcolor{white}
    &  \textbf{LBNet-Ours}           & 735K   &155.8G  &  \textbf{38.13/0.9611}  & \textbf{33.86/0.9194}   & \textbf{32.21/0.9001}    & \textbf{32.58/0.9311}    & \textbf{39.08/0.9780}  \\
    &  \multicolumn{1}{l|}{ {\color[HTML]{369DA2} Gain}}               & \multicolumn{1}{l|}{ {\color[HTML]{369DA2} +4K (0.5\%)}}   &\multicolumn{1}{l|}{ {\color[HTML]{369DA2} +2.6G (1.7\%)}}         &\multicolumn{1}{c|}{{\color[HTML]{369DA2} +0.08/0.0004}}            & \multicolumn{1}{c|}{ {\color[HTML]{369DA2} +0.21/0.0017}}  & \multicolumn{1}{c|}{ {\color[HTML]{369DA2} +0.05/0.0007}}  & \multicolumn{1}{c|}{ {\color[HTML]{369DA2} +0.28/0.0020}}   & \multicolumn{1}{c}{ {\color[HTML]{369DA2} +0.20/0.0005}}    \\

    \hline

     &  EDSR~\cite{lim2017enhanced}               & 1554K   &160.4G & 34.37/0.9270        &30.28/0.8418            & 29.09/0.8052    & 28.15/0.8527   & 33.45/0.9439    \\
    \rowcolor{gray!10}  \cellcolor{white}
    &  \textbf{EDSR-Ours}            & 1568K   &164.7G  &  \textbf{34.52/0.9279}  & \textbf{30.39/0.8424}   & \textbf{29.14/0.8058}    & \textbf{28.36/0.8549}    & \textbf{33.92/0.9460}  \\
    &  \multicolumn{1}{l|}{ {\color[HTML]{369DA2} Gain}}               & \multicolumn{1}{l|}{ {\color[HTML]{369DA2} +14K (0.9\%)}}   &\multicolumn{1}{l|}{ {\color[HTML]{369DA2} +4.3G (2.7\%)}} & \multicolumn{1}{c|}{ {\color[HTML]{369DA2} +0.15/0.0009}}        &\multicolumn{1}{c|}{{\color[HTML]{369DA2} +0.11/0.0006}}            & \multicolumn{1}{c|}{ {\color[HTML]{369DA2} +0.05/0.0006}}    & \multicolumn{1}{c|}{ {\color[HTML]{369DA2} +0.21/0.0022}}   & \multicolumn{1}{c}{ {\color[HTML]{369DA2} +0.47/0.0021}}    \\
    \cmidrule{2-9}
    &  CARN~\cite{ahn2018fast}               & 1592K   &118.8G & 34.29/0.9255        &30.29/0.8407            & 29.06/0.8034    & 28.06/0.8493   & 33.43/0.9427    \\
    \rowcolor{gray!10}  \cellcolor{white}
    $\times 3$ &  \textbf{CARN-Ours}            &1600K     &121.2G  &  \textbf{34.45/0.9277}  & \textbf{30.42/0.8435}   & \textbf{29.12/0.8059}    & \textbf{28.33/0.8547}    & \textbf{33.79/0.9454}  \\
    &  \multicolumn{1}{l|}{ {\color[HTML]{369DA2} Gain}}               & \multicolumn{1}{l|}{ {\color[HTML]{369DA2} +8K (0.5\%)}}   &\multicolumn{1}{l|}{ {\color[HTML]{369DA2} +2.4G (2.0\%)}} & \multicolumn{1}{c|}{ {\color[HTML]{369DA2} +0.16/0.0022}}        &\multicolumn{1}{c|}{{\color[HTML]{369DA2} +0.13/0.0028}}            & \multicolumn{1}{c|}{ {\color[HTML]{369DA2} +0.06/0.0025}}    & \multicolumn{1}{c|}{ {\color[HTML]{369DA2} +0.27/0.0054}}   & \multicolumn{1}{c}{ {\color[HTML]{369DA2} +0.36/0.0027}}    \\
    \cmidrule{2-9}
    &  LBNet~\cite{gao2022lightweight}               & 736K   &68.4G & 34.47/0.9277        &30.38/0.8417            & 29.13/0.8061    & 28.42/0.8559   & 33.82/0.9460    \\
    \rowcolor{gray!10}  \cellcolor{white}
    &  \textbf{LBNet-Ours}             & 740K   &69.4G  &  \textbf{34.55/0.9283}  & \textbf{30.47/0.8438}   & \textbf{29.18/0.8074}    & \textbf{28.66/0.8600}    & \textbf{34.15/0.9477}  \\
    &  \multicolumn{1}{l|}{ {\color[HTML]{369DA2} Gain}}               & \multicolumn{1}{l|}{ {\color[HTML]{369DA2} +4K (0.5\%)}}   &\multicolumn{1}{l|}{ {\color[HTML]{369DA2} +1.0G (1.5\%)}} & \multicolumn{1}{c|}{ {\color[HTML]{369DA2} +0.08/0.0006}}        &\multicolumn{1}{c|}{{\color[HTML]{369DA2} +0.09/0.0021}}            & \multicolumn{1}{c|}{ {\color[HTML]{369DA2} +0.05/0.0013}}    & \multicolumn{1}{c|}{ {\color[HTML]{369DA2} +0.24/0.0041}}   & \multicolumn{1}{c}{ {\color[HTML]{369DA2} +0.33/0.0017}}    \\

    \hline

     &  EDSR~\cite{lim2017enhanced}               &1518K     &114.2G  & 32.09/0.8938        &28.58/0.7813   & 27.57/0.7357    & 26.04/0.7849   & 30.35/0.9067    \\
    \rowcolor{gray!10}  \cellcolor{white}
    &  \textbf{EDSR-Ours}            &1532K     &116.4G  &  \textbf{32.30/0.8960}  &\textbf{28.66/0.7826}   & \textbf{27.63/0.7374}    & \textbf{26.28/0.7899}   & \textbf{30.75/0.9101}   \\
    &  \multicolumn{1}{l|}{ {\color[HTML]{369DA2} Gain}}               & \multicolumn{1}{l|}{ {\color[HTML]{369DA2} +14K (0.9\%)}}   &\multicolumn{1}{l|}{ {\color[HTML]{369DA2} +2.2G (1.9\%)}} & \multicolumn{1}{c|}{ {\color[HTML]{369DA2} +0.21/0.0022}}        &\multicolumn{1}{c|}{{\color[HTML]{369DA2} +0.08/0.0013}}            & \multicolumn{1}{c|}{ {\color[HTML]{369DA2} +0.06/0.0007}}    & \multicolumn{1}{c|}{ {\color[HTML]{369DA2} +0.24/0.0050}}   & \multicolumn{1}{c}{ {\color[HTML]{369DA2} +0.40/0.0034}}    \\
    \cmidrule{2-9}
    &  CARN~\cite{ahn2018fast}               & 1592K  & 90.9G & 32.13/0.8937        & 28.60/0.7806            & 27.58/0.7349    & 26.07/0.7837   & 30.42/0.9070    \\
    \rowcolor{gray!10}  \cellcolor{white}
    $\times 4$
    &  \textbf{CARN-Ours}            &1600K     &92.1G  &  \textbf{32.18/0.8948}  & \textbf{28.64/0.7825}            & \textbf{27.62/0.7369}    & \textbf{26.23/0.7875}   & \textbf{30.70/0.9099}  \\
    &  \multicolumn{1}{l|}{ {\color[HTML]{369DA2} Gain}}               & \multicolumn{1}{l|}{ {\color[HTML]{369DA2} +8K (0.5\%)}}   &\multicolumn{1}{l|}{ {\color[HTML]{369DA2} +1.2G (1.3\%)}} & \multicolumn{1}{c|}{ {\color[HTML]{369DA2} +0.05/0.0011}}        &\multicolumn{1}{c|}{{\color[HTML]{369DA2} +0.04/0.0019}}            & \multicolumn{1}{c|}{ {\color[HTML]{369DA2} +0.04/0.0020}}    & \multicolumn{1}{c|}{ {\color[HTML]{369DA2} +0.16/0.0028}}   & \multicolumn{1}{c}{ {\color[HTML]{369DA2} +0.28/0.0029}}    \\
    \cmidrule{2-9}
    &  LBNet~\cite{gao2022lightweight}               & 742K   &38.9G & 32.29/0.8960        &28.68/0.7832            & 27.62/0.7382    & 26.27/0.7906   & 30.76/0.9111    \\
    \rowcolor{gray!10}  \cellcolor{white}
    &  \textbf{LBNet-Ours}            & 746K   &39.5G  &  \textbf{32.36/0.8967}  & \textbf{28.77/0.7846}   & \textbf{27.69/0.7399}    & \textbf{26.47/0.7947}    & \textbf{31.05/0.9136}  \\
    &  \multicolumn{1}{l|}{ {\color[HTML]{369DA2} Gain}}               & \multicolumn{1}{l|}{ {\color[HTML]{369DA2} +4K (0.5\%)}}   &\multicolumn{1}{l|}{ {\color[HTML]{369DA2} +0.6G (1.5\%)}} & \multicolumn{1}{c|}{ {\color[HTML]{369DA2} +0.07/0.0007}}        &\multicolumn{1}{c|}{{\color[HTML]{369DA2} +0.09/0.0006}}            & \multicolumn{1}{c|}{ {\color[HTML]{369DA2} +0.07/0.0017}}    & \multicolumn{1}{c|}{ {\color[HTML]{369DA2} +0.20/0.0039}}   & \multicolumn{1}{c}{ {\color[HTML]{369DA2} +0.29/0.0025}}    \\

    \bottomrule
\end{tabular}}
\end{table*}

\subsection{Analysis of Strengths of Our FourierSR} \label{subsec:strength}
\subsubsection{Structure Analysis} As shown in Fig.~\ref{fig: difference}, compared with one class of methods represented by GFNet~\cite{rao2021global} and DFFormer~\cite{tatsunami2024fft} in Fig.~\ref{fig: difference} (a), FourierSR shown in Fig.~\ref{fig: difference} (c) additionally does one mix for channel tokens and deals with real and imaginary parts of Fourier features, which allows models to obtain a better representation ability. Specifically, the channel token mix facilitates our model to exchange information in different groups of channels. Processing real and imaginary parts of Fourier features separately is equivalent to data enhancement. Due to the inconsistency of tensor sizes between SR training and testing, global filters can only be defined in the forward process, which seriously reduces model speed. In contrast, using broadcast mechanisms to extend local filters to global filters is more flexible. Another class of methods represented by AFNO~\cite{guibas2021efficient} and AFFNet~\cite{huang2023adaptive} in Fig.~\ref{fig: difference} (b) does too many mixes of channel tokens, resulting in huge costs and slow inference that do not apply to efficient SR. For AFFNet~\cite{huang2023adaptive}, its operation of using a Fourier tensor as a global kernel tends to cause divergence in the training of SR tasks.

\subsubsection{Complexity Comparison} We give a complexity comparison in TABLE~\ref{tab: differnet}. Regarding the number of FLOPs, our method is smaller than the windows-based Transformer (W-Trans), FFC~\cite{chi2020fast}, AFNO~\cite{guibas2021efficient}, and AFFNet~\cite{huang2023adaptive}. The comparison of FourierSR with convolution can be simplified to a comparison of the number of \hbox{${k^2}C$ and ${C \mathord{\left/
 {\vphantom {C \rho }} \right.
 \kern-\nulldelimiterspace} \rho } + 2{\log _2}HW$}. In SR, The height $H$ and width $W$ of the image are typically \hbox{${{1280} \mathord{\left/
 {\vphantom {{1024} s}} \right.
 \kern-\nulldelimiterspace} s}$}, \hbox{${{720} \mathord{\left/
 {\vphantom {{1024} s}} \right.
 \kern-\nulldelimiterspace} s}$}, where $s$ is the scale factor. The kernel size in convolutions is typically $3$. Therefore, our number of FLOPs is smaller than convolutions. Compared to GFNet~\cite{rao2021global}, although we have a slightly higher number of FLOPs, this design contributes to enhancing the model’s representational capacity, including a channel tokens mix and an enhanced convolution range as shown in Fig.~\ref{fig: difference} (c). Moreover, FourierSR has the smallest number of parameters compared to convolutions, window-based Transformer, and existing token mix-based methods. Therefore, we can draw that our FourierSR is a computationally efficient plugin that is superior to common feature extraction modules and token mix methods.

\subsection{Location of Our FourierSR Insertion} \label{subsec:location}
\begin{figure*}[ht]
\begin{overpic}[width=0.99\linewidth]{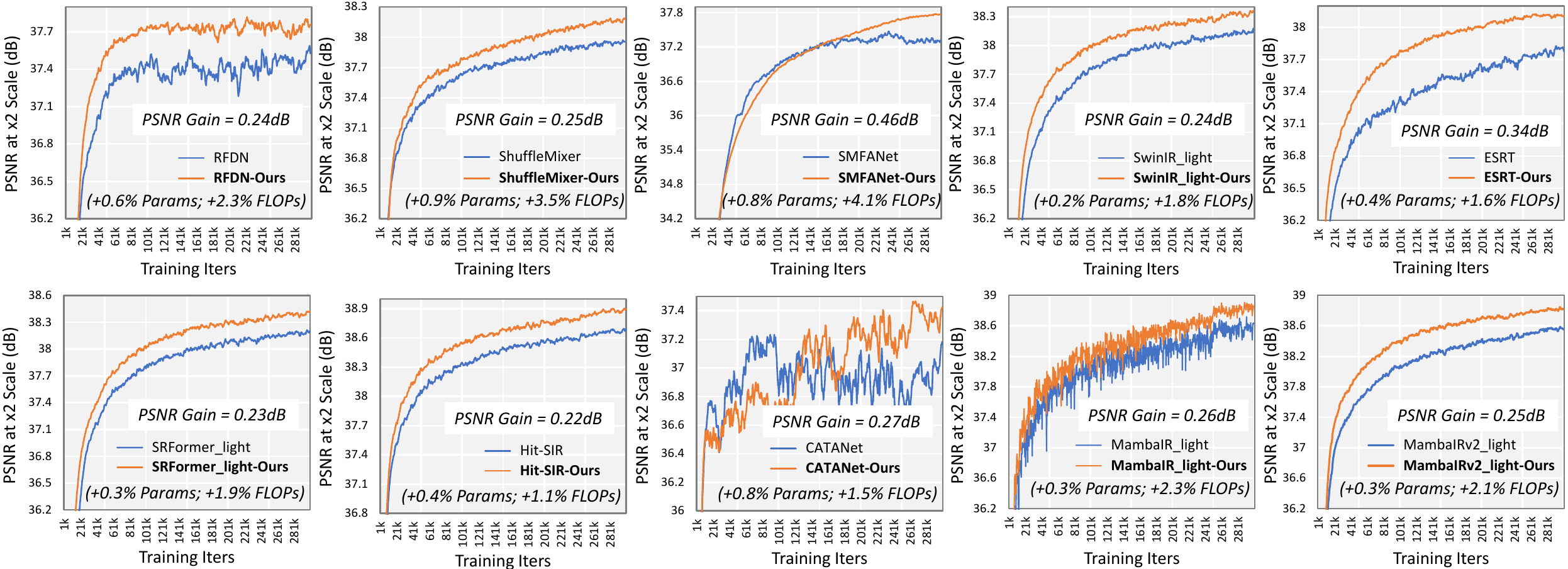}
\end{overpic}
   \caption{Comparison of PSNR in training between existing efficient methods and these methods empowered with FourierSR on Manga109~\cite{matsui2017sketch} test set. We choose CNN-based methods, \eg, RFDN~\cite{liu2020residual}, ShuffleMixer~\cite{sun2022shufflemixer}, and SMFANet~\cite{zheng2025smfanet}, Transformer-based methods, \eg, SwinIR~\cite{liang2021swinir}, ESRT~\cite{lu2022transformer}, SRFormer~\cite{zhou2023srformer}, HiT-SIR~\cite{zhang2024hit}, and CATANet~\cite{liu2025catanet}, and Mamba-based method, \eg, MambaIR~\cite{guo2024mambair} and MambaIRv2~\cite{guo2025mambairv2} as backbones. To speed up training, CNN-based and Mamba-based methods are trained with batch size=8 and patch size=32, and Transformer-based methods are trained with batch size=8 and patch size=16.}
\label{fig: more_psnr_lines}
\vspace{-0mm}
\end{figure*}

To highlight its generality, we randomly insert FourierSR into existing methods. As shown in Fig. \ref{fig: insert_table}, we provide a detailed illustration of the positions where FourierSR is inserted into methods listed in TABLE \ref{tab: performance}. Similarly, as shown in Fig. \ref{fig: insert_figure}, we randomly insert FourierSR into existing methods listed in Fig. \ref{fig: more_psnr_lines}. Overall, we do not intentionally choose the best location for the FourierSR to be inserted to emphasize the plug-and-play nature of FourierSR further. Our FourierSR can improve SR efficiency significantly regardless of where it is inserted and where the method belongs: CNN, Transformer, or Mamba.

 \begin{table}[t!]
\tiny
\setlength\tabcolsep{0.5pt}
\centering
\vspace{0mm}
\caption{Quantitative comparison of existing efficient SR methods empowered with ClassSR~\cite{kong2021classsr} framework and our FourierSR on Test2k~\cite{gu2019div8k}, Test4k~\cite{gu2019div8k}, and Test8K~\cite{gu2019div8k} test sets. Our method reduces complexity by reducing model depths.}
\vspace{-0mm}
\label{tab: Compare_ClassSR}
\resizebox{0.499\textwidth}{!}{
\begin{tabular}{l|l|l|l|c|c|c}
\toprule
& & & & \multicolumn{1}{l|}{Test2K} 
& \multicolumn{1}{l|}{Test4K} 
& \multicolumn{1}{l}{Test8K}  
\\ 
\cmidrule{5-7}
    \multicolumn{1}{l|}{\multirow{-2}{*}{Methods}}  
    & \multicolumn{1}{l|}{\multirow{-2}{*}{Params{\color[HTML]{369DA2} $\downarrow$}}}
    & \multicolumn{1}{l|}{\multirow{-2}{*}{FLOPs{\color[HTML]{369DA2} $\downarrow$}}}
    & \multicolumn{1}{l|}{\multirow{-2}{*}{Speed{\color[HTML]{369DA2} $\downarrow$}}}
    & PSNR{\color[HTML]{369DA2}}
    & PSNR{\color[HTML]{369DA2}}
    & PSNR{\color[HTML]{369DA2}}
    \\ 
    \hline
    FSRCNN~\cite{dong2016accelerating}            &25K     &0.47G  &\textbf{0.9ms} & 25.61       & 26.90           & 32.66     \\
     FSRCNN-ClassSR~\cite{kong2021classsr}            &113K     &0.31G &5.3ms & 25.61       & 26.91           & 32.73     \\
    \rowcolor{gray!10}
     \textbf{FSRCNN-Ours}            &\textbf{23K}     &\textbf{0.30G} &1.6ms & \textbf{25.62}       & \textbf{26.91}           & \textbf{32.74}     \\
     \cmidrule{1-7}
    CARN-M~\cite{ahn2018fast}           &295K     &1.15G   &\textbf{9.1ms}  & 25.95    & 27.34           & 33.18     \\
     CARN-M-ClassSR~\cite{kong2021classsr}            &645K     &0.82G &74.2ms & 26.01       & 27.42           & 33.24     \\
    \rowcolor{gray!10}
     \textbf{CARN-M-Ours}            &\textbf{230K}     &\textbf{0.80G} &11.7ms & \textbf{26.07}      & \textbf{27.47}           & \textbf{33.34}     \\
    \cmidrule{1-7}
     SRResNet~\cite{ledig2017photo}              & 1.5M   &5.20G &11.8ms  & 26.19       & 27.65           & 33.50   \\
      SRResNet-ClassSR~\cite{kong2021classsr}            &3.1M     &3.62G  &42.5ms & 26.20       & 27.66           & 33.50     \\
    \rowcolor{gray!10}
      \textbf{SRResNet-Ours}            &\textbf{1.4M}     &\textbf{3.03G} &\textbf{11.4ms}  & \textbf{26.25}       & \textbf{27.70}           & \textbf{33.57}     \\

    \bottomrule
\end{tabular}}
\end{table}

\begin{figure*}[!t]
	\scriptsize
	\centering
	\scalebox{0.85}{
		\begin{tabular}{lc}
            \begin{adjustbox}{valign=t}
				\begin{tabular}{c}
				\includegraphics[width=0.19\textwidth, height=0.12\textheight]{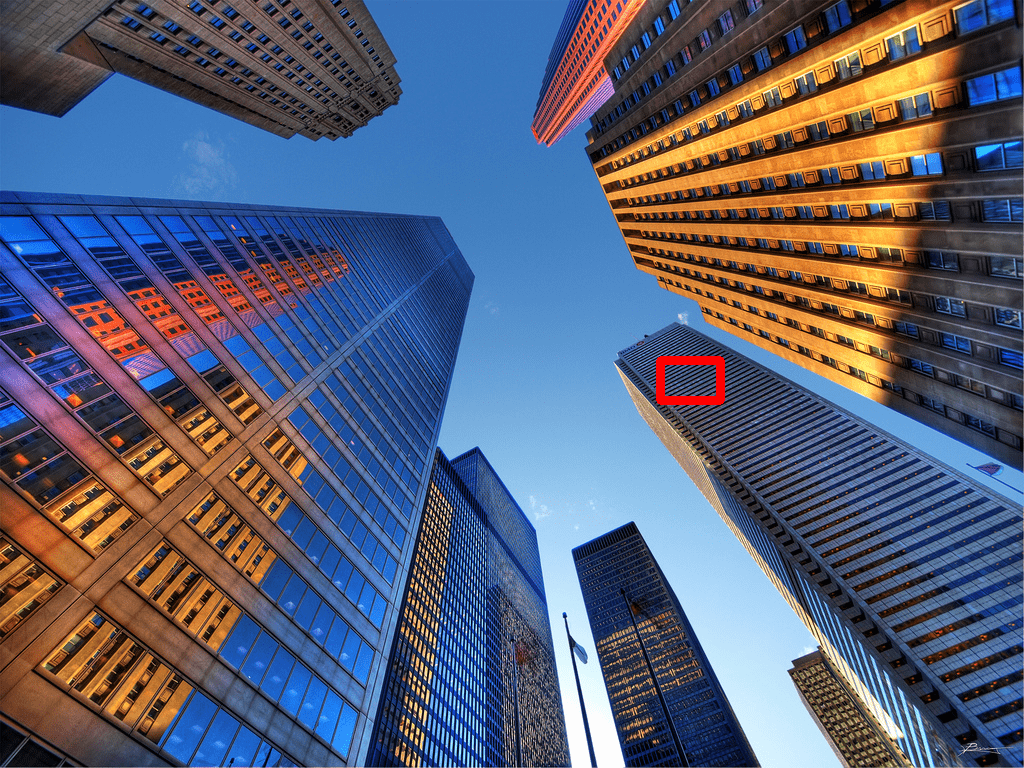} \\
					\fontsize{6.5pt}{1pt}\selectfont Urban100 ($\times 2$): img012 \\
				\end{tabular}
			\end{adjustbox}
			\hspace{-3mm}
			\begin{adjustbox}{valign=t}
				\begin{tabular}{cccc}
					\includegraphics[width=0.08\textwidth, height=0.0525\textheight]{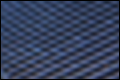} & 
					\hspace{-3mm}
                    \includegraphics[width=0.08\textwidth, height=0.0525\textheight]{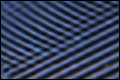} & 
					\hspace{-3mm}
					\includegraphics[width=0.08\textwidth, height=0.0525\textheight]{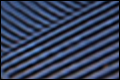}  & 
                    \hspace{-3mm}
					\includegraphics[width=0.08\textwidth, height=0.0525\textheight]{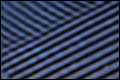} \\
					\fontsize{7pt}{1pt}\selectfont LR & \hspace{-3mm}
					\fontsize{7pt}{1pt}\selectfont EDSR~\cite{lim2017enhanced} & \hspace{-3mm}
                    \fontsize{7pt}{1pt}\selectfont CARN~\cite{ahn2018fast} & 
                    \hspace{-3mm}
                    \fontsize{7pt}{1pt}\selectfont LBNet~\cite{gao2022lightweight}\\
					
					\includegraphics[width=0.08\textwidth, height=0.0525\textheight]{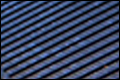} & 
					\hspace{-3mm}
                    \includegraphics[width=0.08\textwidth, height=0.0525\textheight]{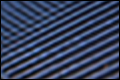} & 
					\hspace{-3mm}
					\includegraphics[width=0.08\textwidth, height=0.0525\textheight]{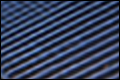}  & 
                    \hspace{-3mm}
					\includegraphics[width=0.08\textwidth, height=0.0525\textheight]{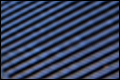} \\
					\fontsize{7pt}{1pt}\selectfont Ground Truth & \hspace{-3mm}
                    \fontsize{7pt}{1pt}\selectfont \textbf{EDSR-Ours} & \hspace{-3mm}
					\fontsize{7pt}{1pt}\selectfont \textbf{CARN-Ours} &  \hspace{-3mm}
					\fontsize{7pt}{1pt}\selectfont \textbf{LBNet-Ours}\\
				\end{tabular}
			\end{adjustbox}

                        \begin{adjustbox}{valign=t}
				\begin{tabular}{c}
				\includegraphics[width=0.19\textwidth, height=0.12\textheight]{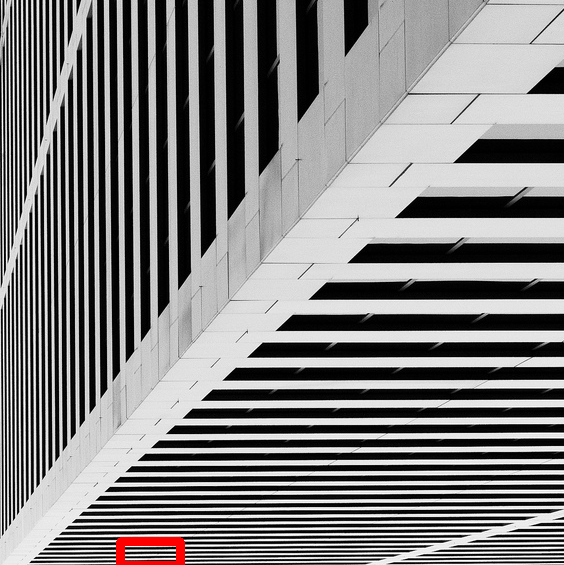} \\
					\fontsize{6.5pt}{1pt}\selectfont Urban100 ($\times 2$): img011 \\
				\end{tabular}
			\end{adjustbox}
			\hspace{-3mm}
			\begin{adjustbox}{valign=t}
				\begin{tabular}{cccc}
					\includegraphics[width=0.08\textwidth, height=0.0525\textheight]{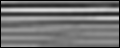} & 
					\hspace{-3mm}
                    \includegraphics[width=0.08\textwidth, height=0.0525\textheight]{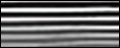} & 
					\hspace{-3mm}
					\includegraphics[width=0.08\textwidth, height=0.0525\textheight]{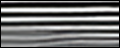}  & 
                    \hspace{-3mm}
					\includegraphics[width=0.08\textwidth, height=0.0525\textheight]{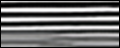} \\
					\fontsize{7pt}{1pt}\selectfont LR & \hspace{-3mm}
					\fontsize{7pt}{1pt}\selectfont EDSR~\cite{lim2017enhanced} & \hspace{-3mm}
                    \fontsize{7pt}{1pt}\selectfont CARN~\cite{ahn2018fast} & 
                    \hspace{-3mm}
                    \fontsize{7pt}{1pt}\selectfont LBNet~\cite{gao2022lightweight}\\
					
					\includegraphics[width=0.08\textwidth, height=0.0525\textheight]{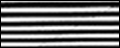} & 
					\hspace{-3mm}
                    \includegraphics[width=0.08\textwidth, height=0.0525\textheight]{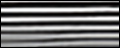} & 
					\hspace{-3mm}
					\includegraphics[width=0.08\textwidth, height=0.0525\textheight]{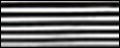}  & 
                    \hspace{-3mm}
					\includegraphics[width=0.08\textwidth, height=0.0525\textheight]{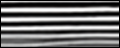} \\
					\fontsize{7pt}{1pt}\selectfont Ground Truth & \hspace{-3mm}
                    \fontsize{7pt}{1pt}\selectfont \textbf{EDSR-Ours} & \hspace{-3mm}
					\fontsize{7pt}{1pt}\selectfont \textbf{CARN-Ours} &  \hspace{-3mm}
					\fontsize{7pt}{1pt}\selectfont \textbf{LBNet-Ours}\\
				\end{tabular}
			\end{adjustbox}

            			\\

            \begin{adjustbox}{valign=t}
				\begin{tabular}{c}
				\includegraphics[width=0.19\textwidth, height=0.12\textheight]{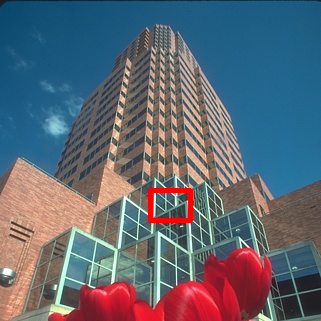} \\
					\fontsize{6.5pt}{1pt}\selectfont B100 ($\times 3$): 86000 \\
				\end{tabular}
			\end{adjustbox}
			\hspace{-3mm}
			\begin{adjustbox}{valign=t}
				\begin{tabular}{cccc}
					\includegraphics[width=0.08\textwidth, height=0.0525\textheight]{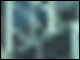} & 
					\hspace{-3mm}
                    \includegraphics[width=0.08\textwidth, height=0.0525\textheight]{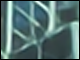} & 
					\hspace{-3mm}
					\includegraphics[width=0.08\textwidth, height=0.0525\textheight]{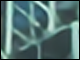}  & 
                    \hspace{-3mm}
					\includegraphics[width=0.08\textwidth, height=0.0525\textheight]{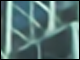} \\
					\fontsize{7pt}{1pt}\selectfont LR & \hspace{-3mm}
					\fontsize{7pt}{1pt}\selectfont EDSR~\cite{lim2017enhanced} & \hspace{-3mm}
                    \fontsize{7pt}{1pt}\selectfont CARN~\cite{ahn2018fast} & 
                    \hspace{-3mm}
                    \fontsize{7pt}{1pt}\selectfont LBNet~\cite{gao2022lightweight}\\
					
					\includegraphics[width=0.08\textwidth, height=0.0525\textheight]{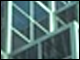} & 
					\hspace{-3mm}
                    \includegraphics[width=0.08\textwidth, height=0.0525\textheight]{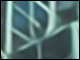} & 
					\hspace{-3mm}
					\includegraphics[width=0.08\textwidth, height=0.0525\textheight]{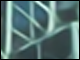}  & 
                    \hspace{-3mm}
					\includegraphics[width=0.08\textwidth, height=0.0525\textheight]{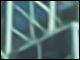} \\
					\fontsize{7pt}{1pt}\selectfont Ground Truth & \hspace{-3mm}
                    \fontsize{7pt}{1pt}\selectfont \textbf{EDSR-Ours} & \hspace{-3mm}
					\fontsize{7pt}{1pt}\selectfont \textbf{CARN-Ours} &  \hspace{-3mm}
					\fontsize{7pt}{1pt}\selectfont \textbf{LBNet-Ours}\\
				\end{tabular}
			\end{adjustbox}

            \begin{adjustbox}{valign=t}
				\begin{tabular}{c}
				\includegraphics[width=0.19\textwidth, height=0.12\textheight]{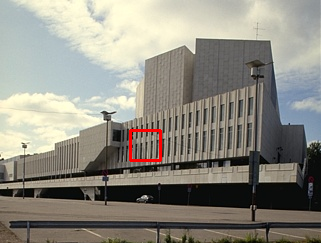} \\
					\fontsize{6.5pt}{1pt}\selectfont BSDS100 ($\times 3$): 78004 \\
				\end{tabular}
			\end{adjustbox}
			\hspace{-3mm}
			\begin{adjustbox}{valign=t}
				\begin{tabular}{cccc}
					\includegraphics[width=0.08\textwidth, height=0.0525\textheight]{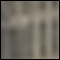} & 
					\hspace{-3mm}
                    \includegraphics[width=0.08\textwidth, height=0.0525\textheight]{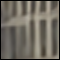} & 
					\hspace{-3mm}
					\includegraphics[width=0.08\textwidth, height=0.0525\textheight]{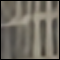}  & 
                    \hspace{-3mm}
					\includegraphics[width=0.08\textwidth, height=0.0525\textheight]{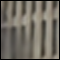} \\
					\fontsize{7pt}{1pt}\selectfont LR & \hspace{-3mm}
					\fontsize{7pt}{1pt}\selectfont EDSR~\cite{lim2017enhanced} & \hspace{-3mm}
                    \fontsize{7pt}{1pt}\selectfont CARN~\cite{ahn2018fast} & 
                    \hspace{-3mm}
                    \fontsize{7pt}{1pt}\selectfont LBNet~\cite{gao2022lightweight}\\
					
					\includegraphics[width=0.08\textwidth, height=0.0525\textheight]{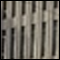} & 
					\hspace{-3mm}
                    \includegraphics[width=0.08\textwidth, height=0.0525\textheight]{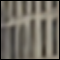} & 
					\hspace{-3mm}
					\includegraphics[width=0.08\textwidth, height=0.0525\textheight]{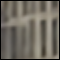}  & 
                    \hspace{-3mm}
					\includegraphics[width=0.08\textwidth, height=0.0525\textheight]{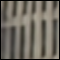} \\
					\fontsize{7pt}{1pt}\selectfont Ground Truth & \hspace{-3mm}
                    \fontsize{7pt}{1pt}\selectfont \textbf{EDSR-Ours} & \hspace{-3mm}
					\fontsize{7pt}{1pt}\selectfont \textbf{CARN-Ours} &  \hspace{-3mm}
					\fontsize{7pt}{1pt}\selectfont \textbf{LBNet-Ours}\\
				\end{tabular}
			\end{adjustbox}

            \\

            \begin{adjustbox}{valign=t}
				\begin{tabular}{c}
				\includegraphics[width=0.19\textwidth, height=0.12\textheight]{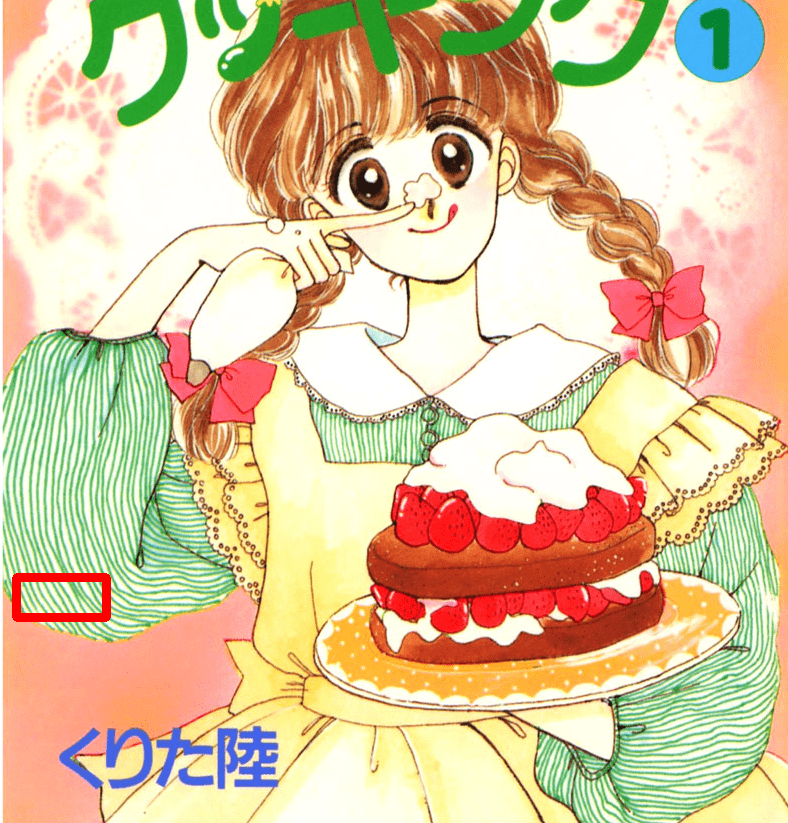} \\
					\fontsize{6.5pt}{1pt}\selectfont Manga109 ($\times 4$): YumeiroCooking \\
				\end{tabular}
			\end{adjustbox}
			\hspace{-3mm}
			\begin{adjustbox}{valign=t}
				\begin{tabular}{cccc}
					\includegraphics[width=0.08\textwidth, height=0.0525\textheight]{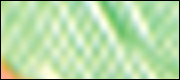} & 
					\hspace{-3mm}
                    \includegraphics[width=0.08\textwidth, height=0.0525\textheight]{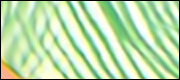} & 
					\hspace{-3mm}
					\includegraphics[width=0.08\textwidth, height=0.0525\textheight]{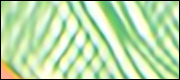}  & 
                    \hspace{-3mm}
					\includegraphics[width=0.08\textwidth, height=0.0525\textheight]{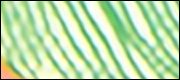} \\
					\fontsize{7pt}{1pt}\selectfont LR & \hspace{-3mm}
					\fontsize{7pt}{1pt}\selectfont EDSR~\cite{lim2017enhanced} & \hspace{-3mm}
                    \fontsize{7pt}{1pt}\selectfont CARN~\cite{ahn2018fast} & 
                    \hspace{-3mm}
                    \fontsize{7pt}{1pt}\selectfont LBNet~\cite{gao2022lightweight}\\
					
					\includegraphics[width=0.08\textwidth, height=0.0525\textheight]{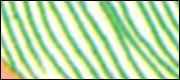} & 
					\hspace{-3mm}
                    \includegraphics[width=0.08\textwidth, height=0.0525\textheight]{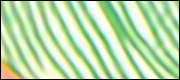} & 
					\hspace{-3mm}
					\includegraphics[width=0.08\textwidth, height=0.0525\textheight]{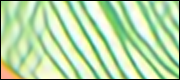}  & 
                    \hspace{-3mm}
					\includegraphics[width=0.08\textwidth, height=0.0525\textheight]{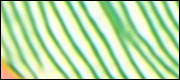} \\
					\fontsize{7pt}{1pt}\selectfont Ground Truth & \hspace{-3mm}
                    \fontsize{7pt}{1pt}\selectfont \textbf{EDSR-Ours} & \hspace{-3mm}
					\fontsize{7pt}{1pt}\selectfont \textbf{CARN-Ours} &  \hspace{-3mm}
					\fontsize{7pt}{1pt}\selectfont \textbf{LBNet-Ours}\\
				\end{tabular}
			\end{adjustbox}

            \begin{adjustbox}{valign=t}
				\begin{tabular}{c}
				\includegraphics[width=0.19\textwidth, height=0.12\textheight]{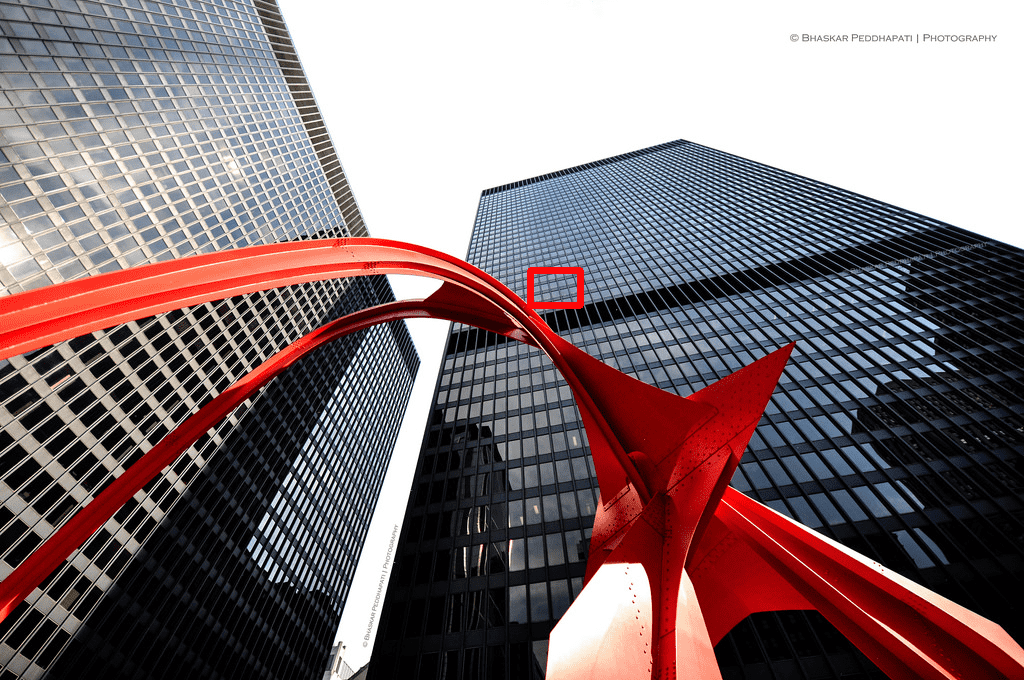} \\
					\fontsize{6.5pt}{1pt}\selectfont Urban100 ($\times 4$): img062 \\
				\end{tabular}
			\end{adjustbox}
			\hspace{-3mm}
			\begin{adjustbox}{valign=t}
				\begin{tabular}{cccc}
					\includegraphics[width=0.08\textwidth, height=0.0525\textheight]{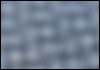} & 
					\hspace{-3mm}
                    \includegraphics[width=0.08\textwidth, height=0.0525\textheight]{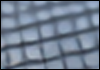} & 
					\hspace{-3mm}
					\includegraphics[width=0.08\textwidth, height=0.0525\textheight]{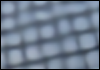}  & 
                    \hspace{-3mm}
					\includegraphics[width=0.08\textwidth, height=0.0525\textheight]{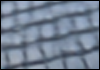} \\
					\fontsize{7pt}{1pt}\selectfont LR & \hspace{-3mm}
					\fontsize{7pt}{1pt}\selectfont EDSR~\cite{lim2017enhanced} & \hspace{-3mm}
                    \fontsize{7pt}{1pt}\selectfont CARN~\cite{ahn2018fast} & 
                    \hspace{-3mm}
                    \fontsize{7pt}{1pt}\selectfont LBNet~\cite{gao2022lightweight}\\
					
					\includegraphics[width=0.08\textwidth, height=0.0525\textheight]{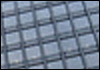} & 
					\hspace{-3mm}
                    \includegraphics[width=0.08\textwidth, height=0.0525\textheight]{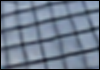} & 
					\hspace{-3mm}
					\includegraphics[width=0.08\textwidth, height=0.0525\textheight]{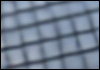}  & 
                    \hspace{-3mm}
					\includegraphics[width=0.08\textwidth, height=0.0525\textheight]{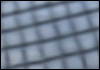} \\
					\fontsize{7pt}{1pt}\selectfont Ground Truth & \hspace{-3mm}
                    \fontsize{7pt}{1pt}\selectfont \textbf{EDSR-Ours} & \hspace{-3mm}
					\fontsize{7pt}{1pt}\selectfont \textbf{CARN-Ours} &  \hspace{-3mm}
					\fontsize{7pt}{1pt}\selectfont \textbf{LBNet-Ours}\\
				\end{tabular}
			\end{adjustbox}
   
			\\
			
	\end{tabular} }
	\vspace{-0mm}
	\caption{Qualitative comparisons between existing efficient SR methods and these methods empowered with our FourierSR.}
	\label{fig: visual_benchmarks}
 \vspace{-0mm}
\end{figure*}

\section{Experiments}

\subsection{Datasets and Metrics} We use the first 800 images of DIV2K~\cite{timofte2017ntire} as our training data and evaluate methods on five benchmarks, including Set5~\cite{bevilacqua2012low}, Set14~\cite{zeyde2012single}, BSDS100~\cite{martin2001database}, Urban100~\cite{huang2015single}, and Manga109~\cite{matsui2017sketch}. To compare with ClassSR \cite{kong2021classsr} fairly, we deal with DIV2K like ClassSR \cite{kong2021classsr} to generate HR and LR training images. Meanwhile, we deal with DIV8K~\cite{gu2019div8k} to generate Test2K-8K test sets. As for evaluation metrics, we calculate PSNR and SSIM~\cite{wang2002universal} on the Y channel in the YCbCr colour.

\begin{table*}[hbtp]
\tiny
\setlength\tabcolsep{2pt}
\centering
\vspace{0mm}
\caption{Quantitative evaluation of a larger size version of existing methods listed in TABLE~\ref{tab: performance} and these methods empowered with our FourierSR. Our FourierSR demonstrates outstanding efficiency when applied in existing methods.}
\vspace{-0mm}
\label{tab: compare_L}
\resizebox{0.98\textwidth}{!}{
\begin{tabular}{c|l|l|l|l|c|c|c|c|c}
\toprule
& & & & & \multicolumn{1}{c|}{Set5 \cite{bevilacqua2012low}} 
& \multicolumn{1}{c|}{Set14 \cite{zeyde2012single}} 
& \multicolumn{1}{c|}{BSDS100 \cite{martin2001database}} 
& \multicolumn{1}{c|}{Urban100 \cite{huang2015single}} 
& \multicolumn{1}{c}{Manga109 \cite{matsui2017sketch}} 
\\ 
\cmidrule{6-10}
    \multicolumn{1}{c|}{\multirow{-2}{*}{Scale}}
    & \multicolumn{1}{l|}{\multirow{-2}{*}{Methods}}  
    & \multicolumn{1}{l|}{\multirow{-2}{*}{Params{\color[HTML]{369DA2} $\downarrow$}}}
    & \multicolumn{1}{l|}{\multirow{-2}{*}{FLOPs{\color[HTML]{369DA2} $\downarrow$}}}
    & \multicolumn{1}{l|}{\multirow{-2}{*}{Speed{\color[HTML]{369DA2} $\downarrow$}}}
    & PSNR{\color[HTML]{369DA2}}/SSIM{\color[HTML]{369DA2}}
    & PSNR{\color[HTML]{369DA2}}/SSIM{\color[HTML]{369DA2}}
    & PSNR{\color[HTML]{369DA2}}/SSIM{\color[HTML]{369DA2}}
    & PSNR{\color[HTML]{369DA2}}/SSIM{\color[HTML]{369DA2}}  
    & PSNR{\color[HTML]{369DA2}}/SSIM{\color[HTML]{369DA2}}
    \\ 
    \hline
     &  EDSR-L~\cite{lim2017enhanced}               & 2552K   &588.1G  & \textbf{52ms} & 38.03/0.9606        &33.62/0.9177            & 32.18/0.8997    & 32.17/0.9288   & 38.69/0.9771    \\
    \rowcolor{gray!10}
    \cellcolor{white} &  \textbf{EDSR-Ours}            & \textbf{1384K}   &\textbf{326.5G}    & 58ms &  \textbf{38.07/0.9608}  & \textbf{33.66/0.9182}   & \textbf{32.19/0.8998}    & \textbf{32.27/0.9290}    & \textbf{38.89/0.9775}  \\
    \cmidrule{2-10}
    $\times 2$ &  CARN-L~\cite{ahn2018fast}               & 2368K   &401.2G   & 51ms & 37.83/0.9591       & 33.54/0.9168           & 32.10/0.8978    & 31.95/0.9259     & 38.45/0.9770  \\
    \rowcolor{gray!10}  \cellcolor{white}
    &  \textbf{CARN-Ours}            &\textbf{1600K}     &\textbf{228.6G}    & \textbf{49ms}  & \textbf{38.03/0.9605}       & \textbf{33.65/0.9179}           & \textbf{32.15/0.8993}    & \textbf{32.10/0.9276}     & \textbf{38.74/0.9771}  \\
    \cmidrule{2-10}
    &  LBNet-L~\cite{gao2022lightweight}               & 923K   &193.7G   & 1465ms & 38.07/0.9608       & 33.68/0.9180           & 32.17/0.8994    & 32.50/0.9301     & 38.97/0.9777    \\
    \rowcolor{gray!10}  \cellcolor{white}
    &  \textbf{LBNet-Ours}           & \textbf{735K}   &\textbf{155.8G}   & \textbf{1204ms} &  \textbf{38.13/0.9611}  & \textbf{33.86/0.9194}   & \textbf{32.21/0.9001}    & \textbf{32.58/0.9311}    & \textbf{39.08/0.9780}  \\

    \bottomrule
\end{tabular}}
\end{table*}

\begin{figure}[t]
\hspace{0mm}
\vspace{-1mm}
\begin{overpic}[width=0.99\linewidth]{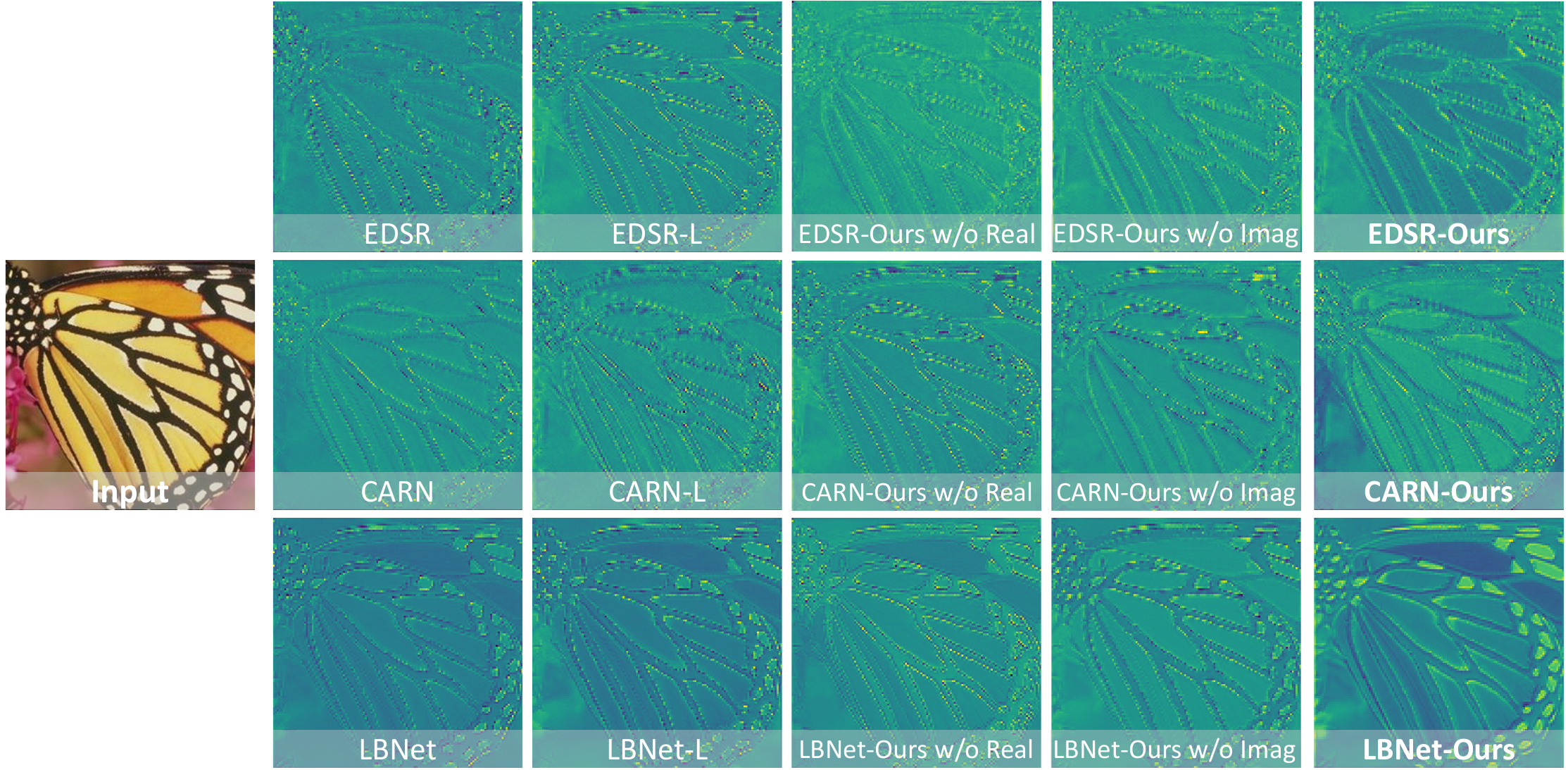}
\end{overpic}
\vspace{0mm}
   \caption{Visualization of feature maps at the same depth across different models, including a large-size version of existing methods and the upper (Real) and lower (Imag) branches of FourierSR. Our FourierSR enhances the model's ability to capture comprehensive features.}
\label{fig: feature_maps}
\vspace{-0mm}
\end{figure}


\subsection{Implementation Details}
We use the PyTorch framework with an NVIDIA GeForce RTX 4090 for all of the experiments. The learning rate, patch size, batch size, and optimizer settings are identical to the original methods. In the TABLE~\ref{tab: performance}, the number of our FourierSR is 16, 9, and 9 in EDSR~\cite{lim2017enhanced}, CARN~\cite{ahn2018fast}, and LBNet~\cite{gao2022lightweight}, respectively. In the Fig.~\ref{fig: more_psnr_lines}, the number of our FourierSR is 4, 5, 4, 6, 6, 6, 4, 8, 4 and 4 in RFDN~\cite{liu2020residual}, ShuffleMixer~\cite{sun2022shufflemixer}, SMFANet~\cite{zheng2025smfanet}, SwinIR~\cite{liang2021swinir}, ESRT~\cite{lu2022transformer}, SRFormer~\cite{zhou2023srformer}, Hit-SIR~\cite{zhang2024hit}, CATANet~\cite{liu2025catanet}, MambaIR~\cite{guo2024mambair}, and MambaIRv2~\cite{guo2025mambairv2}. The location where our FourierSR inserts into the existing methods network is random. Since FourierSR involves the computation of Fourier real and imaginary, to minimize errors that this part may cause, we introduce our Fourier Token Loss for supervision in our Supplementary Material. FLOPs and speed are measured corresponding to an HR image of the spatial size of 1280 × 720 pixels. 


\subsection{Benchmarks Evaluation}
We choose existing efficient SR methods, including CNN-based methods, \eg, EDSR~\cite{lim2017enhanced}, CARN~\cite{ahn2018fast}, and Transformer-based methods, \eg, LBNet~\cite{gao2022lightweight}, as benchmarks to evaluate our FourierSR's effectiveness. As shown in TABLE~\ref{tab: performance}, for existing methods, our FourierSR as a plug-and-play brings an average PSNR gain of 0.18 dB in five test sets, the average increase of Params and FLOPs counts of 0.6\% and 2\% of the original size. Furthermore, we compare the visual results of original benchmarks and benchmarks plus our FourierSR in Fig.~\ref{fig: visual_benchmarks}. Due to the improved global ability of models after embedding our FourierSR, existing methods can handle a larger range of features, as well as more subtle features. As a result, existing methods plus our FourierSR facilitate the correction of incorrectly recovered textures and the elimination of artifacts introduced by reconstruction. Since our method enhances the model's ability to capture features by expanding its receptive field. Textures, due to their higher structures, benefit more from this enhancement and are thus more visibly improved. In contrast, other areas like details, may not show as pronounced a change because they tend to have tinier structures.


To further validate the generalization of our FourierSR, we test our FourierSR on more efficient SR methods, including CNN-based methods, \eg, RFDN~\cite{liu2020residual}, ShuffleMix~\cite{sun2022shufflemixer}, SMFANet~\cite{zheng2025smfanet}, Transformer-based methods, \eg, SwinIR~\cite{liang2021swinir}, ESRT~\cite{lu2022transformer}, SRFormer~\cite{zhou2023srformer}, HiT-SIR~\cite{zhang2024hit}, CATANet~\cite{liu2025catanet}, and Mamba-based method, \eg, MambaIR~\cite{guo2024mambair}, MambaIRv2~\cite{guo2025mambairv2} to further show FourierSR's generalization. Due to GPU resource constraints, we train CNN-based and Mamba-based methods with batch size=8 and patch size=32 and Transformer-based methods with batch size=8 and patch size=16 to speed up training. As shown in Fig.~\ref{fig: more_psnr_lines}, whether original methods are CNN-based, Transformer-based, or Mamba-based, FourierSR can be a plugin to significantly improve their performance with a tiny increase in computational costs. It is worth noting that for methods ShuffleMix~\cite{sun2022shufflemixer} and SMFANet~\cite{zheng2025smfanet}, which use extra Fourier-based loss, our FourierSR is equally effective in improving their performance by 0.25 dB and 0.46 dB, respectively. Evaluation of 13 efficient SR methods fully illustrates the generalizability of our FourierSR. 


\begin{figure}[t]
\vspace{-1mm}
\begin{overpic}[width=0.98\linewidth]{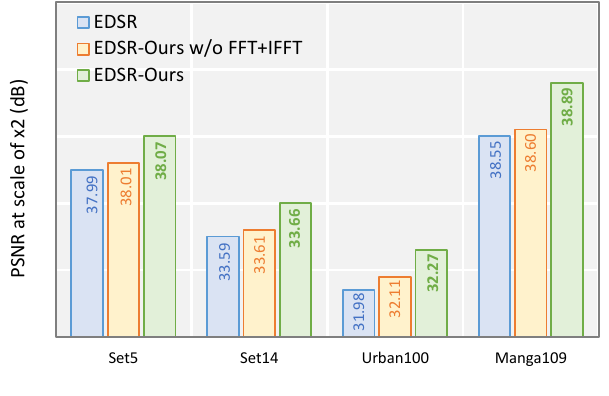}
\end{overpic}
\vspace{-3mm}
   \caption{Ablation study on the theoretical validity of our FourierSR, where the convolution theorem no longer holds when Fourier transformers (FFT+IFFT) are not applied.}
\label{fig: theoretical}
\vspace{-0mm}
\end{figure}

\begin{table*}[t!]
\tiny
\setlength\tabcolsep{2.5pt}
\centering
\vspace{0mm}
    \caption{Ablation study on components of FourierSR shown in Fig.~\ref{fig: main} on Manga109~\cite{matsui2017sketch} test set at a scale of $\times$2.}
\vspace{-0mm}
\label{tab: FourierSR}
\resizebox{0.99\textwidth}{!}{
\begin{tabular}{l|ccccc|l|l|cc}
\toprule
    \multicolumn{1}{l|}{\multirow{-1}{*}{Methods}} 
    & Channel Tokens Mix  & Real Part    
    & Imag Part  & Upper Branch     
    & Lower Branch
    & \multicolumn{1}{c|}{\multirow{-1}{*}{Params{\color[HTML]{369DA2} $\downarrow$}}}
    & \multicolumn{1}{c|}{\multirow{-1}{*}{FLOPs{\color[HTML]{369DA2} $\downarrow$}}}
    & PSNR  & SSIM 
    \\ 
    \hline
    \emph{w/o} CTM  & \XSolidBrush   & \textcolor{gray}{\Checkmark} & \textcolor{gray}{\Checkmark} & \textcolor{gray}{\Checkmark}     & \textcolor{gray}{\Checkmark}   & 1378K  & 324.6G  & 38.72\textsubscript{\color[HTML]{369DA2}  \scalebox{0.75}{-0.17}} & 0.9772\textsubscript{\color[HTML]{369DA2}  \scalebox{0.75}{-0.0003}}  \\
    \emph{w/o} Real  & \textcolor{gray}{\Checkmark} & \XSolidBrush  & \textcolor{gray}{\Checkmark}  & \textcolor{gray}{\Checkmark}     & \textcolor{gray}{\Checkmark}    & 1382K  & 326.5G  & 38.75\textsubscript{\color[HTML]{369DA2}  \scalebox{0.75}{-0.14}}  & 0.9771\textsubscript{\color[HTML]{369DA2}  \scalebox{0.75}{-0.0004}}  \\
    \emph{w/o} Imag  & \textcolor{gray}{\Checkmark} & \textcolor{gray}{\Checkmark}  & \XSolidBrush  & \textcolor{gray}{\Checkmark}     & \textcolor{gray}{\Checkmark}      & 1382K  & 326.5G  & 38.77\textsubscript{\color[HTML]{369DA2}  \scalebox{0.75}{-0.12}}  & 0.9771\textsubscript{\color[HTML]{369DA2}  \scalebox{0.75}{-0.0004}}  \\
    \emph{w/o} Upper  &\textcolor{gray}{\Checkmark}  & \textcolor{gray}{\Checkmark}  & \textcolor{gray}{\Checkmark}  & \XSolidBrush     & \textcolor{gray}{\Checkmark}     & 1382K  & 326.5G  & 38.78\textsubscript{\color[HTML]{369DA2}  \scalebox{0.75}{-0.11}}  & 0.9772\textsubscript{\color[HTML]{369DA2}  \scalebox{0.75}{-0.0003}}  \\
    \emph{w/o} Lower  &\textcolor{gray}{\Checkmark}  &\textcolor{gray}{\Checkmark}   &\textcolor{gray}{\Checkmark}   & \textcolor{gray}{\Checkmark} & \XSolidBrush     & 1382K  & 326.5G  & 38.76\textsubscript{\color[HTML]{369DA2}  \scalebox{0.75}{-0.13}}  & 0.9771\textsubscript{\color[HTML]{369DA2}  \scalebox{0.75}{-0.0004}}  \\
    \rowcolor{gray!10}
    \textbf{EDSR-Ours}  & \textcolor{gray}{\Checkmark}  & \textcolor{gray}{\Checkmark} & \textcolor{gray}{\Checkmark}  & \textcolor{gray}{\Checkmark}  & \textcolor{gray}{\Checkmark}     & 1384K  & 326.5G  & \textbf{38.89}  & \textbf{0.9775}  \\
    \bottomrule
\end{tabular}}
\end{table*}
\hspace{-0mm}

\subsubsection{Further Reduce Network Complexity}
Although our FourierSR expand the receptive field of models to improve SR, it still comes at the cost of a small increase in computational cost. To reduce the model computational cost, we can reduce the width or depth of existing methods to control network complexity further. In TABLE~\ref{tab: Compare_ClassSR}, we compare our FourierSR with a model accelerated framework called ClassSR~\cite{kong2021classsr}. For a fair comparison, our training and testing setting is the same as ClassSR framework. Our FourierSR performs better than ClassSR in terms of the number of Params and FLOPs, inference speed, and PSNR values. Specifically, our method improves SR performance while significantly reducing the computational costs of original SR methods. Additionally, it's worth noting that ClassSR requires an increase in the number of Params of original methods by a factor of $2$ to $4$ to reduce the FLOPs count. Therefore, ClassSR slows down model inference speed. In contrast, inserting our FourierSR allows methods to reduce model depth or width, thus reducing computational costs and speeding up model inference. Therefore, we can draw that our FourierSR provide a solution to further reduce network complexity while maintain the SR performance.

\subsection{Ablation Study}
\subsubsection{Theoretical Validity}
We demonstrate whether the theory of FourierSR holds in two cases. \textbf{First:} Is the PSNR gain due to increased computational costs? As shown in TABLE~\ref{tab: compare_L}, we compare existing methods in TABLE~\ref{tab: performance} inserted with our FourierSR and a larger size version of these methods, where FourierSR performs much better in PSNR values due to its ability to capture global receptive fields. Additionally, due to the extremely lightweight nature of our FourierSR, it has a much lower number of parameters and FLOPs than the large-size version of these methods at a reasonable inference speed. Next, we show a visualization of feature maps at the same depth under different models in Fig.~\ref{fig: feature_maps}. Models can enhance the sharpness of high-frequency regions in feature maps by increasing computational costs, but the improvement is limited. In contrast, our FourierSR is more effective in recovering a clear high-frequency feature map. \textbf{Second:} Is the PSNR gain due to the convolution theorem? As shown in Fig.~\ref{fig: theoretical}, we remove the Fourier transform (FFT) and inverse Fourier transform (IFFT) to disallow the convolution theorem, resulting in extremely limited PSNR gains caused by the inability to model global convolution. These two parts of ablation studies illustrate that the improved performance is not due to the increased computational cost of our FourierSR but rather to the successful simulation of global convolution.

\begin{table}[t]
\tiny
\setlength\tabcolsep{1pt}
\centering
\caption{Ablation study on the impact of the number of channel token mixes (CTM) in our FourierSR.}
\vspace{-0mm}
\label{tab: number_CTM}
\resizebox{0.48\textwidth}{!}{
\begin{tabular}{c|l|l|l|c}
\toprule
& & & & \multicolumn{1}{c}{Manga109~\cite{matsui2017sketch}} 
\\ 
\cmidrule{5-5}
    \multicolumn{1}{c|}{\multirow{-2}{*}{Scale}}
    & \multicolumn{1}{l|}{\multirow{-2}{*}{Methods}} 
    & \multicolumn{1}{c|}{\multirow{-2}{*}{Params{\color[HTML]{369DA2} $\downarrow$}}}
    & \multicolumn{1}{c|}{\multirow{-2}{*}{FLOPs{\color[HTML]{369DA2} $\downarrow$}}}
    & PSNR/SSIM \\ 
    \hline
    \multirow{5}{*}{$\times 2$}  &EDSR-Ours (0CTM)   & 1373K & 324.6G & 38.72/0.9772\\ 
    \rowcolor{gray!10} \cellcolor{white}
    & \textbf{EDSR-Ours (1CTM)}   & 1384K & 326.5G & 38.89\textsubscript{\color[HTML]{369DA2}  \scalebox{0.8}{+0.17}}/0.9775\\ 
    & EDSR-Ours (2CTM)  & 1392K & 328.4G & 38.92\textsubscript{\color[HTML]{369DA2}  \scalebox{0.8}{+0.20}}/0.9775\\  
    & EDSR-Ours (4CTM)   & 1409K & 332.2G & 38.94\textsubscript{\color[HTML]{369DA2}  \scalebox{0.8}{+0.22}}/0.9775\\ 
    & EDSR-Ours (8CTM)   & 1442K & 339.4G & 38.90\textsubscript{\color[HTML]{369DA2}  \scalebox{0.8}{+0.18}}/0.9774\\

    \bottomrule
\end{tabular}}
\vspace{-0mm}
\end{table}

\subsubsection{Modular Components}
As shown in TABLE~\ref{tab: FourierSR}, we verify the effectiveness of modular components in our FourierSR. First, we show the effectiveness of the channel token mix (CTM) in FourierSR in the first row, in which CTM greatly enhances the model representation by exchanging information on different channel tokens, achieving a PSNR gain of 0.17dB at the cost of increasing the number of parameters by only 4K and FLOPs by only 2.1G. Additionally, as shown in TABLE~\ref{tab: number_CTM}, we explain why we use one channel token mix (CTM) in our FourierSR. This is because the performance gain is insignificant when using more than $1$ CTM in FourierSR. From the perspective of model efficiency, we insert $1$ CTM in FourierSR from an efficiency point of view.

\begin{table}[t]
\tiny
\setlength\tabcolsep{1pt}
\centering
\caption{Comparison of our FourierSR with \textbf{(Top)} Convolutional Neural Network (CNN)~\cite{zheng2025smfanet}, 
windows-based Transformer (WTrans)~\cite{liu2025catanet}, and \textbf{(Bottom)} existing token mix-based methods~\cite{rao2021global,chi2020fast,guibas2021efficient,huang2023adaptive,tatsunami2024fft,lou2025transxnet}.}
\label{tab: compare_CT}
\resizebox{0.48\textwidth}{!}{
\begin{tabular}{c|l|l|l|l|c}
\toprule
& & & & & \multicolumn{1}{c}{Manga109~\cite{matsui2017sketch}} 
\\ 
\cmidrule{6-6}
     \multicolumn{1}{c|}{\multirow{-2}{*}{Scale}}
    & \multicolumn{1}{l|}{\multirow{-2}{*}{Methods}} 
    & \multicolumn{1}{l|}{\multirow{-2}{*}{Speed{\color[HTML]{369DA2} $\downarrow$}}} 
    & \multicolumn{1}{c|}{\multirow{-2}{*}{Params{\color[HTML]{369DA2} $\downarrow$}}}
    & \multicolumn{1}{c|}{\multirow{-2}{*}{FLOPs{\color[HTML]{369DA2} $\downarrow$}}}
    & PSNR/SSIM\\ 
    \hline
    \multirow{10}{*}{$\times 2$}  & 
    EDSR  & 30ms & 1370K & 316.2G & 38.55/0.9769\\ 
    & EDSR-CNN~\cite{zheng2025smfanet}  & 81ms & 2063K & 456.6G & 38.68/0.9772\\ 
    & EDSR-WTrans~\cite{liu2025catanet}  & 162ms & 1819K & 404.7G & 38.72/0.9773 \\
    \cmidrule{2-6}
    & EDSR-GFNet~\cite{rao2021global}  & 155ms & 30.8M & 326.6G & 38.70/0.9771\\ 
    & EDSR-AFNO~\cite{guibas2021efficient}  & 134ms & 1472K & 329.4G & 38.83/0.9773\\ 
    & EDSR-FFC~\cite{chi2020fast}  & 76ms & 1502K & 363.2G & 38.65/0.9771\\ 
    & EDSR-AFFNet~\cite{huang2023adaptive}  & 146ms & 1574K & 342.6G & NAN/NAN \\
    & EDSR-FourierGNN~\cite{yi2024fouriergnn}  & 181ms  & 1523K & 336.1G & 38.72/0.9772 \\
    & EDSR-DMixer~\cite{lou2025transxnet}  & 122ms & 1622K & 386.1G & 38.64/0.9770 \\
    
    \rowcolor{gray!10} \cellcolor{white}
    & \textbf{EDSR-Ours}  & \textbf{58ms} & \textbf{1384K} & \textbf{326.5G} & \textbf{38.89/0.9775}\\

    \bottomrule
\end{tabular}}
\vspace{-0mm}
\end{table}

Then, we explain the importance of utilizing the Fourier features' real and imaginary parts separately in FourierSR, which has been proven in reasoning that can enhance the global receptive field of models. As shown in the second and third rows of TABLE~\ref{tab: FourierSR}, this operator leads to an average PSNR increase of 0.13 dB while only growing the number of Params by 2K. Furthermore, as shown in Fig.~\ref{fig: feature_maps}, we visualize the upper and lower branches of FourierSR separately. Removing the upper branch (real part) reduces large-scale, low-frequency structures, while removing the lower branch (imag part) diminishes high-frequency details and edges, resulting in a less clear feature map. Next, we verify the effectiveness of adding and subbing the upper and lower branches to expand the receptive field theoretically. Similarly, as shown in the fourth and fifth rows of TABLE~\ref{tab: FourierSR}, this operator increases the PSNR by an average of 0.12 dB with only a 2K increase in the number of parameters, which fully illustrates the lightweight of such an operation and its ability to enhance global receptive fields effectively. 


\subsubsection{Rationality of Designs}
\textbf{First}, as shown in the top of TABLE~\ref{tab: compare_CT}, we compare our FourierSR with CNN~\cite{zheng2025smfanet} and Transformer~\cite{liu2025catanet}, where CNN has limited performance gains, and the Transformer severely impacts model inference speeds. In contrast, our FourierSR is a better feature-extracting unit, which can further improve SR performance with fewer Params and FLOPs counts at a suitable inference speed. \textbf{Second}, as shown in the bottom of TABLE~\ref{tab: compare_CT}, we show that our FourierSR design is more suitable for efficient SR than existing Fourier token-based methods. GFNet~\cite{rao2021global} defines global filters of the same size as the feature map as convolution kernels lead to huge parameter counts. The use of excessive matrix multiplications and built-in convolutions in AFNO~\cite{guibas2021efficient}, FFC~\cite{chi2020fast}, FourierGNN~\cite{yi2024fouriergnn}, and DMixer~\cite{lou2025transxnet} significantly impacts the model's inference speed and computational cost. AFFNet~\cite{huang2023adaptive} defines Fourier features as global convolution kernels, leading to gradient explosion in SR. In contrast, our FourierSR improves SR the most and incurs a small cost.

\begin{table}[t]
\tiny
\setlength\tabcolsep{1pt}
\centering
\caption{Comparison of the inference speed, Params counts, and FLOPs counts for a $3\times3$ convolution, a windows-based Transformer, and a FourierSR, where the convolutional kernel size is 3 and the number of channels is 64.}
\vspace{-0mm}
\label{tab: latency}
\resizebox{0.48\textwidth}{!}{
\begin{tabular}{c|c|c|c|c}
\toprule
& & & & 
\\ 
    \multicolumn{1}{c|}{\multirow{-2}{*}{Scale}}
    & \multicolumn{1}{l|}{\multirow{-2}{*}{Modules}} 
    & \multicolumn{1}{l|}{\multirow{-2}{*}{$3\times3$ Conv}} 
    & \multicolumn{1}{c|}{\multirow{-2}{*}{windows-based Trans}}
    & \multicolumn{1}{c}{\multirow{-2}{*}{\textbf{FourierSR}}}
     \\ 
    \hline
    \multirow{3}{*}{$\times 2$}  & 
    Latency{\color[HTML]{369DA2} $\downarrow$}  & \textbf{0.4ms} & 517ms & 2.0ms \\ 
    & 
    Params{\color[HTML]{369DA2} $\downarrow$}  & 36.864K & 16.384K & \textbf{0.896K} \\ 
    & 
    FLOPs{\color[HTML]{369DA2} $\downarrow$}  & 8.493G & 5.662G & \textbf{0.644G} \\ 

    \bottomrule
\end{tabular}}
\vspace{-0mm}
\end{table}

\section{Limitation and Future Work}
Since the presence of the Fast Fourier Transform (FFT) and Inverse Fast Fourier Transform (IFFT) within our FourierSR affects the model's inference speed to a certain extent, our FourierSR is suited to be used as a plug-in to incorporate into existing methods under certain quantitative constraints, rather than on a large scale. This ensures that methods empowered by our FourierSR maintain a reasonable inference speed while improving the model representation.

Specifically, comparing the $3\times3$ convolution and windows-based Transformer, the strength of our FourierSR lies in its lightweight nature with powerful representation capabilities, as can be seen from Param's counts and FLOPs counts comparisons in TABLE~\ref{tab: latency}, and performance comparisons in TABLE~\ref{tab: compare_CT}. However, according to the latency comparison of TABLE~\ref{tab: latency}, although our FourierSR has a huge advantage in inference speed compared to the windows-based Transformer, the inference speed advantage over the $3\times3$ convolution is insignificant. Therefore, when FourierSR is used on a large scale as a feature enhancement module in existing methods, inference speed will be slower than CNN-based models. 

Inspired by sparse-based and window-based works that aim to reduce complexity, we will leverage sparse representations and window divisions in the future to further reduce the complexity of FFT and IFFT, thereby accelerating the inference of FourierSR while maintaining the ability to boost SR. Furthermore, the all-in-one restoration~\cite{jiang2025survey} builds a unified framework capable of handling multiple low-level tasks, including adverse weather restoration~\cite{sun2024restoring,li2026seeing}, SR~\cite{li2023cross}, and deblur~\cite{kong2023efficient}. Our FourierSR exhibits potential owing to its plug-and-play nature, which we plan to further explore.

\section{Conclusion}
This paper proposes FourierSR, a plugin to uniformly improve off-the-shelf efficient SR methods with low computational costs and global receptive fields. To achieve this goal, motivated by the token mix and the convolution theorem, real or imaginary parts of the Fourier domain features perform the token mix with global filters obtained by expanding our defined local filters. We demonstrate that the above operations are mathematically equivalent to performing global convolutions on the input plus the inverse of the input, avoiding the instability or inefficiency in existing token mixes. Experiments show that incorporating FourierSR into existing methods can significantly extend its receptive field, significantly improving efficient SR performance with a minimal computational cost.

\bibliographystyle{IEEEtran}
\bibliography{sample-base}

\end{document}